%% file: main.tex
\documentclass{article} % For LaTeX2e
\usepackage{iclr2026_conference,times}

\input{math_commands.tex}

\usepackage{hyperref}
\usepackage{url}

\usepackage[utf8]{inputenc} % allow utf-8 input
\usepackage[T1]{fontenc}    % use 8-bit T1 fonts
\usepackage{hyperref}       % hyperlinks
\usepackage{url}            % simple URL typesetting
\usepackage{booktabs}       % professional-quality tables
\usepackage{amsfonts}       % blackboard math symbols
\usepackage{nicefrac}       % compact symbols for 1/2, etc.
\usepackage{microtype}      % microtypography
\usepackage[dvipsnames]{xcolor}

\usepackage{algorithm} 
\usepackage{algorithmicx}
\usepackage{algpseudocode}

\usepackage{hyperref}
\usepackage{url}
\usepackage[utf8]{inputenc} % allow utf-8 input
\usepackage[T1]{fontenc}    % use 8-bit T1 fonts
\usepackage{nicefrac}       % compact symbols for 1/2, etc.
\usepackage{multirow}
\usepackage{subcaption}
\usepackage{xspace}
\usepackage{amsmath}
\usepackage{svg}
\usepackage{bbm}
\usepackage{dsfont}
\usepackage{graphicx}
\usepackage{siunitx}
\usepackage{kotex}
\usepackage{xcolor}
\usepackage[table]{xcolor}% http://ctan.org/pkg/xcolor

\newcommand{\ie}{\textit{i.e.\xspace}}
\newcommand{\eg}{\textit{e.g.\xspace}}

\newcommand{\data}{\textsc{GeoFact-X}\xspace}
\newcommand{\model}{\textsc{M2A}\xspace}

\title{Learn Globally, Speak Locally:\\Bridging the Gaps in Multilingual Reasoning}

\author{%
Jaedong Hwang$^1$\thanks{Equal Contribution \quad $^\dagger$Corresponding Authors}\qquad
Kumar Tanmay$^{2*}$\qquad
Seok-Jin Lee$^{3*}$\qquad
Ayush Agrawal$^{4,5}$\\
\textbf{Hamid Palangi$^6$\qquad
Kumar Ayush$^{6,7}$\qquad
Ila Fiete$^{1\dagger}$\qquad
Paul Pu Liang$^{1\dagger}$} \vspace{0.1cm} \\
$^1$Massachusetts Institute of Technology\quad
$^2$Havard Univeristy\quad
$^3$LG CNS\\
$^4$Université de Montréal\quad
$^5$Mila\quad
$^6$Google\quad
$^7$Stanford University\\
}

\iclrfinalcopy % Uncomment for camera-ready version, but NOT for submission.
\begin{document}

\maketitle

%%%%%%%%% ABSTRACT
\begin{abstract}
   \input{sections/0_abstract}

\end{abstract}

\newenvironment{Itemize}{
    \begin{itemize}[leftmargin=*]
    \setlength{\itemsep}{0pt}
    \setlength{\topsep}{0pt}
    \setlength{\partopsep}{0pt}
    \setlength{\parskip}{1pt}}
{\end{itemize}}
\setlength{\leftmargini}{9pt}

\input{sections/1_introduction}
\input{sections/2_related}
\input{sections/3_method}
\input{sections/4_benchmark}

\input{sections/5_revisit}
\input{sections/6_exp}
\input{sections/7_conclusion}

\section*{Ethics and Reproducible Statement}
%We read the ICLR Code of Ethics before the submission.
Our paper focuses on multilingual reasoning capabilities of large language models (LLMs), emphasizing knowledge transfer between languages and highlighting limitations faced by low-resource languages.
We believe this work encourages the research community to address these limitations, ultimately contributing toward equitable access to high-performing LLMs, regardless of the user's language.
However, our work also shares similar negative societal concerns with standard large language model research (\eg, biased toward high-resource languages and hallucination).
We specify the experimental setting in Appendix~\ref{supp:exp_details} and attached the codebase as supplementary materials.

%\section*{Reproducible Statement}

\bibliography{egbib,continual_learning}
\bibliographystyle{iclr2026_conference}

\appendix
\input{sections/8_appendix}

\section{LLM Usage}
We used LLM to help with paper writing for improving grammar and wording.

\end{document}

%% file: math_commands.tex
\usepackage{amsmath,amsfonts,bm}

\def\eqref#1{equation~\ref{#1}}
\def\1{\bm{1}}

\DeclareMathAlphabet{\mathsfit}{\encodingdefault}{\sfdefault}{m}{sl}
\SetMathAlphabet{\mathsfit}{bold}{\encodingdefault}{\sfdefault}{bx}{n}

%% file: sections/0_abstract.tex
%!TEX root = ./../main.tex

Large Language Models (LLMs) have achieved strong performance in domains like mathematics, factual question answering, and code generation, yet their ability to reason on these tasks in different languages remains underdeveloped.
Especially for low-resource languages such as Swahili or Thai, LLMs can often misinterpret prompts or default to reasoning in English.
This implicit bias toward high-resource languages undermines factual accuracy, interpretability, and trust.
We propose \model, a novel method that combines multi-scale multilingual alignment with language-consistency rewards on machine-translated questions, training models to reason directly and accurately in the target language.
Furthermore, existing multilingual benchmarks only evaluate on final answers, overlooking whether reasoning occurs in the intended language.
To close this gap, we introduce \data, a geography-based multilingual factual reasoning benchmark together with reasoning traces in five languages: English, Hindi, Japanese, Swahili, and Thai.
Our results show that \model significantly enhances multilingual reasoning fidelity in both mathematical and factual reasoning tasks,  highlighting that reasoning-aware multilingual reinforcement learning is crucial for robust cross-lingual generalization\footnote{\url{https://jd730.github.io/projects/M2A_GeoFact-X}}.

%% file: sections/1_introduction.tex
%!TEX root = ./../main.tex

\section{Introduction}
\label{sec:intro}

Large Language Models (LLMs) have made remarkable progress in reasoning tasks, such as mathematics~\citep{liu2024your,shao2024deepseekmath}, code generation~\citep{jain2024livecodebench,team2025gemma}, and factual QA~\citep{achiam2023gpt,guo2025deepseek,qwen2025qwen2.5,wang2024mmlu}, primarily in English.
Yet, their reasoning capabilities remain underdeveloped in low-resource languages such as Swahili, Marathi, or Thai~\citep{cahyawijaya2024llms,nguyen2023democratizing}.
This performance disparity undermines the \textit{trustworthiness} of LLMs in these languages since users cannot check the reasoning traces to verify the answer.
For this purpose, both the final answer and the intermediate reasoning should ideally be expressed in the question language to ensure \textit{interpretability}, \ie, users can directly follow the reasoning in their own language.
The central issue is not only whether LLMs can provide correct answers, but whether they can \emph{think} in the language of the question. When they cannot, translation of reasoning traces offers only a partial solution and one that often fails on cultural and linguistic nuance.
Recent studies~\citep{aggarwal2025language,yao2024benchmarking} suggest that both LLMs and machine translation systems struggle with cultural and linguistic nuances.
For example, culturally grounded concepts such as Chinese \textit{Guānxi} (关系), Japanese \textit{Wa} (和), and Korean \textit{Jeong} (정)  remain difficult to capture faithfully.

In this work, we conduct the first comprehensive study of \textbf{multilingual reasoning}: assessing whether LLMs can not only answer questions correctly, but also  \textit{reason in the same language as the question}.
Prior multilingual benchmarks primarily assess the final accuracy~\citep{ponti2020xcopa,shi2023language,xuan2025mmluprox}, overlooking the language of the reasoning traces.
By evaluating reasoning traces directly on MGSM~\citep{shi2023language}, we found that models often revert to English reasoning even under non-English prompts.
This gap between the prompt language and the reasoning process underscores a broader problem that LLMs can appear correct while failing to reason in a language. Ensuring both accuracy and language-consistent reasoning is essential for globally inclusive and interpretable AI.

To tackle this challenge, we introduce \model (Multi-Scale Multilingual Alignment), an efficient approach that explicitly enforces language-consistent reasoning while retaining factual correctness.
The key idea is to combine multi-scale multilingual reasoning alignment with a language-consistency reward, providing reinforcement-learning signals that encourage reasoning traces to remain in the question language, enabling reasoning capabilities to be learned without ground-truth supervision in that language.
We jointly employ supervised fine-tuning and group relative policy optimization (GRPO) to integrate supervised learning on ground-truth reasoning traces with reinforcement-based refinement.
Unlike prior work~\citep{guo2025deepseek,liu2025understanding,ranaldi2025multilingual} that optimizes only for correctness or formality, our method targets the \emph{alignment of reasoning itself}.

Despite recent progress in multilingual LLMs~\citep{ahuja2023mega,qin2025survey}, their ability to perform factual reasoning across cultural contexts remains largely unevaluated. 
We introduce \data, a benchmark of culturally grounded questions localized to five countries (USA, India, Japan, Kenya, Thailand) in their predominant languages (English, Hindi, Japanese, Swahili, Thai). 
By grounding evaluation in country-specific knowledge, \data enables systematic assessment of whether LLMs can reason faithfully within linguistically and culturally contextualized spaces.

\begin{figure}[t!]
    \centering
    \includegraphics[width=\textwidth]{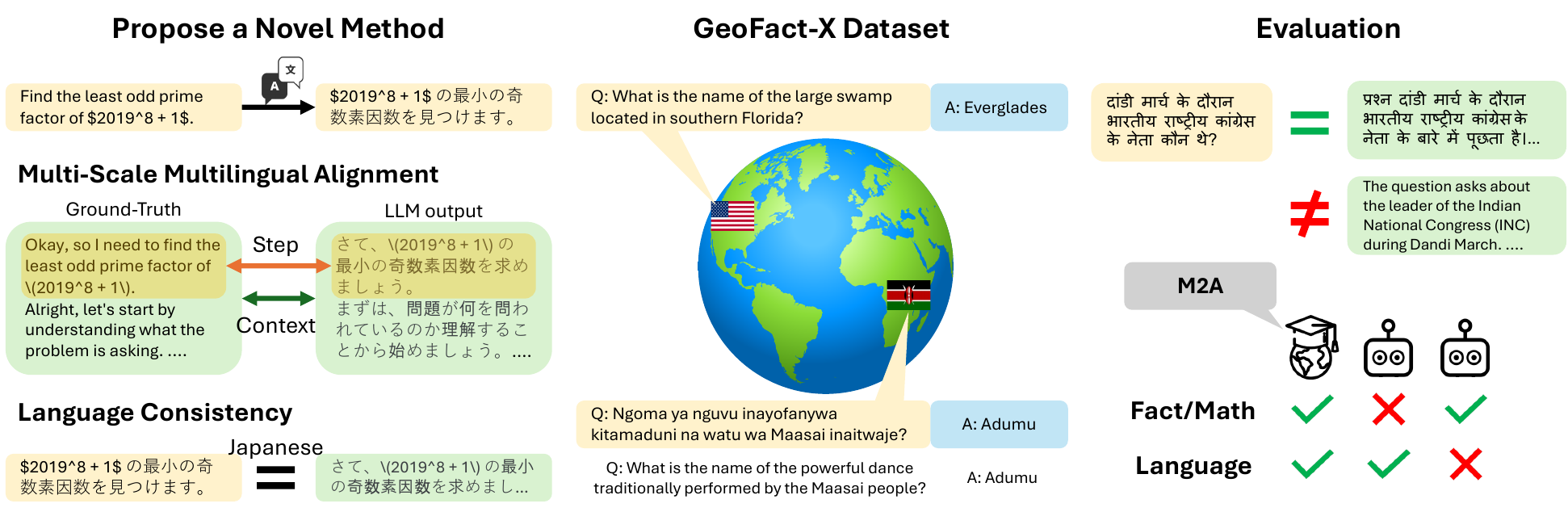}
    \caption{
    \textbf{Illustration of Contributions}
    We propose \model, a new method that utilizes multi-scale multilingual alignment and language consistency rewards from a given machine-translated question, enabling reasoning in the question language.
    We also introduce \data, a new multilingual factual reasoning benchmark which includes training datasets and step-by-step reasoning traces across five languages.
    We propose an automatic evaluation protocol to assess whether a model reasons in the question language and the correctness of reasoning via language identifier or LLM-as-a-judge. 
    }
    \label{fig:concept}
\end{figure}

We train \model on the s1K-1.1~\citep{muennighoff2023crosslingual} dataset and \data train set for mathematical and factual reasoning, respectively.
Our experiments demonstrate that \model yields significant improvement on multilingual reasoning in both cases while reasoning in the question languages.
Figure~\ref{fig:concept} summarizes our key contributions.
Together with the release of our code, data, and evaluation protocols, our work provides a foundation for future work on multilingual reasoning.

%% file: sections/2_related.tex
%!TEX root = ./../main.tex
\section{Related Work}
\label{sec:related}

\subsection{Multilingualism in Large Language Models}
Recent advances in multilingual large language models (LLMs) mark a shift from monolingual dominance to more inclusive cross-lingual capabilities.
Two perspectives frame much of this discourse.
The scaling view holds that increasing the volume and diversity of multilingual data during pretraining enhances cross-lingual generalization~\citep{xue2021mt5,conneau2020unsupervised,chang2023multilinguality,gurgurov2024multilingual}, though it faces the \textit{curse of multilinguality}, where accommodating many languages dilutes performance in individual ones due to limited parameter capacity.
The optimization view emphasizes careful fine-tuning as a means of preserving and amplifying multilingual knowledge~\citep{devlin2019bert,conneau2020unsupervised, luo2023empirical,zhai2023investigating}, yet aggressive post-training risks catastrophic forgetting by overwriting deeply embedded linguistic priors.
Beyond scaling and optimization, \citet{schut2025do} and \citet{zhong2025language} analyzed the internal representations of multilingual models, while \citet{yong2025crosslingual} proposed test-time strategies for improving cross-lingual reasoning.
Notably, \citet{schut2025do} found that multilingual LLMs often perform intermediate reasoning in English, but their scope was limited to internal representation analysis rather than generated reasoning traces.
More recently, post-training alignment techniques such as Direct Preference Optimization (DPO)~\citep{rafailov2023direct} have been applied to embed multilingual reasoning~\citep{dang2024rlhf,ranaldi2025multilingual}.  
Similarly, we employ Group Relative Policy Optimization (GRPO)~\citep{guo2025deepseek,shao2024deepseekmath} with multi-scale multilingual alignment and language consistency rewards, leveraging explicit reasoning traces in the original languages to build stronger multilingual reasoning capabilities.

\subsection{Evaluation of Multilingual Reasoning Capabilities}
Instruction-tuning datasets such as Bactrian~\citep{li2023bactrian}, Aya~\citep{singh2024aya}, Multilingual Alpaca~\citep{chen2023monolingual}, and SphinX~\citep{ahuja2024sphinx} have improved performance across high- and low-resource languages by emphasizing diversity and cultural specificity.
Complementary to these efforts, benchmarks like MEGA~\citep{ahuja2023mega} provide broad multilingual coverage across 70 languages and 16 NLP tasks, but primarily evaluate task-level accuracy rather than reasoning processes.
Other multilingual reasoning benchmarks, including XCOPA~\citep{ponti2020xcopa}, XWinograd~\citep{tikhonov2021s}, and XStoryCloze~\citep{lin2022few}, adopt multiple-choice formats that permit shallow guessing and suffer from translation artifacts~\citep{li2024can}.
In contrast, our benchmark directly targets multilingual reasoning by evaluating free-form, step-by-step generation with explicit reasoning traces.
This design enables more faithful assessment of both reasoning quality and language alignment, providing a sharper diagnostic tool for multilingual LLMs.

%% file: sections/3_method.tex
%!TEX root = ./../main.tex

\begin{figure}[t!]
    \centering
    \includegraphics[width=\textwidth]{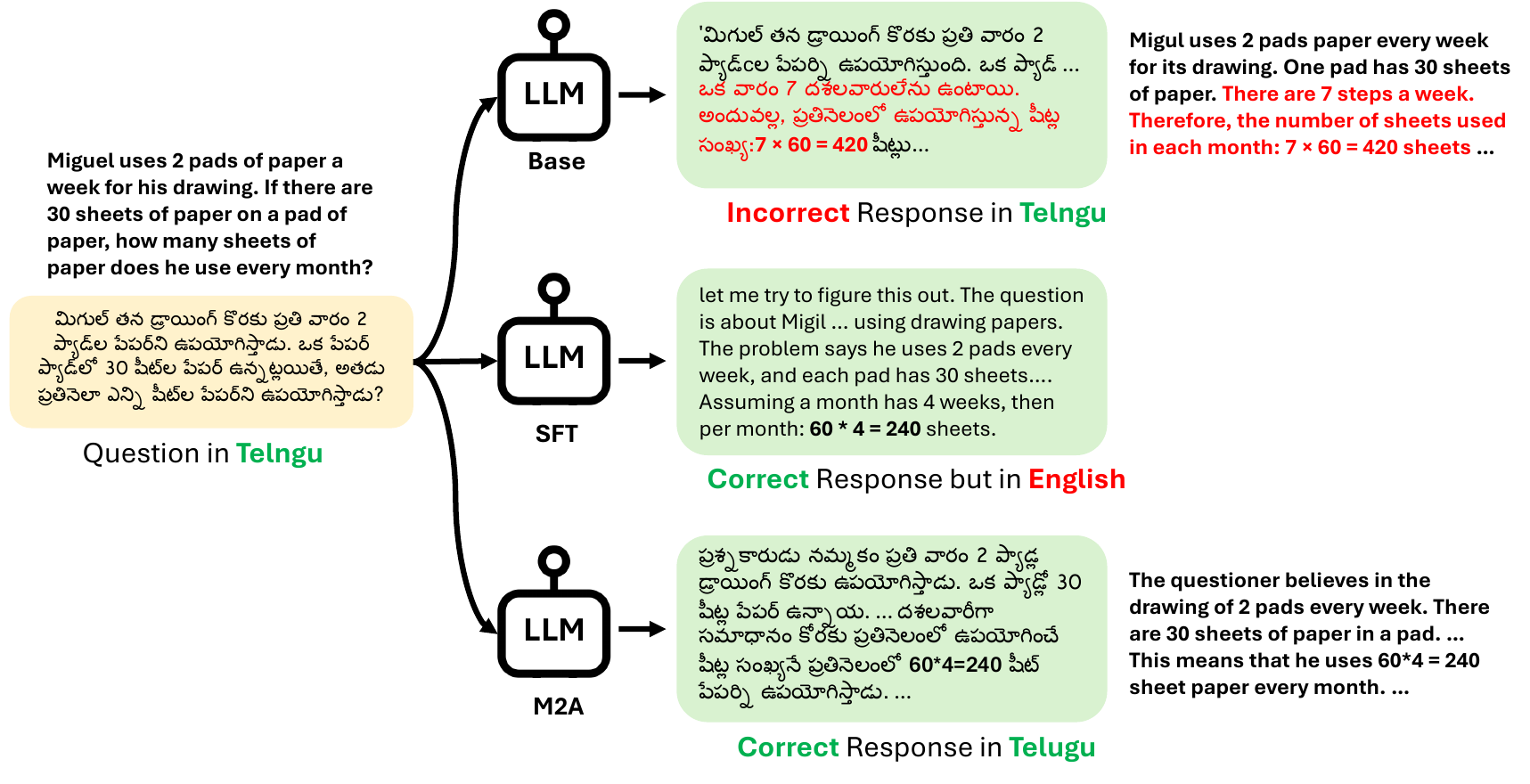}
    \caption{
    \textbf{Example of outputs from a Telugu question from three different models.}
    Base LLM and supervised fine-tuned models (SFT) are correct only in either answer or language, whereas our \model is correct in both answer and language.
    }
    \label{fig:method}
\end{figure}

\section{\model: Multi-Scale Multilingual Alignment}
\label{sec:method}

We propose a new method to enrich existing English-based reasoning models with multilingual reasoning capabilities.
Our approach combines the complementary strengths of supervised fine-tuning~(SFT) and reinforcement learning~(RL).
While SFT enables base reasoning capabilities in English, we find that these models struggle with multilingual reasoning during test time as shown in Figure~\ref{fig:method}.
To that end, we propose a new test-time RL method for multilingual reasoning.
Key to this new approach is defining the right set of rewards that incentivize the model to reason consistently across different languages.
We first translate each question into multiple languages by using Google Translate, allowing the model to generate outputs conditioned on the translated inputs.
We then define a set of multi-scale rewards across different multilingual granularities.
A context alignment reward measures multilingual alignment across the entire reasoning context with the original ground-truth reasoning trace.
This is followed by a reasoning-step alignment reward that aligns each individual reasoning step to capture fine-grained correspondence.
Finally, a language consistency reward explicitly enforces reasoning in the question language.

\begin{figure}[t!]
    \centering
    \includegraphics[width=\textwidth]{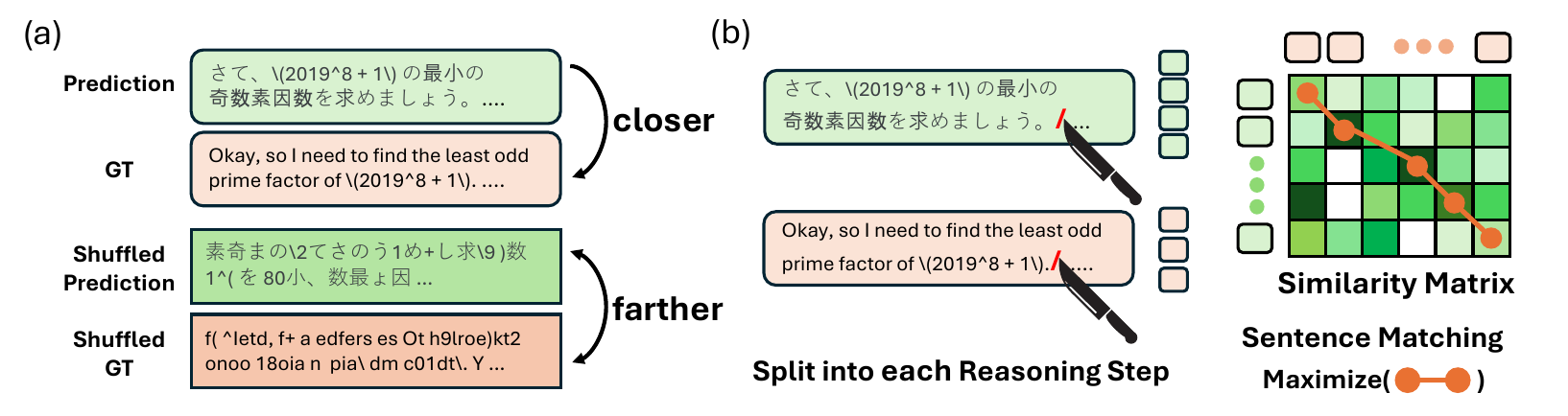}
    \caption{
    \textbf{Overview of \model.}
(a) \emph{Multilingual Context Alignment} enforces global similarity between generated and reference responses while discouraging trivial matches via shuffled negatives. 
(b) \emph{Multilingual Reasoning-Step Alignment} provides finer-grained supervision by aligning individual reasoning steps with ground-truth traces using dynamic programming.
    }
    \label{fig:m2a}
\end{figure}

\paragraph{Problem Setup.} 
Given the question sampled from the question dataset, $q \sim P(\mathcal{Q})$ and its corresponding ground-truth response $y$, we translate each question to the target language $l$, $q'$ by using a machine translator (\eg, Google Translate).
GRPO generates a group of outputs $\{o_1, o_2, ..., o_G\}$ from the translated question.
The reward $r_t$ is calculated based on each output $o_t$.

\paragraph{Multilingual Context Alignment.}
We first encode both output and the ground-truth with the encoder, $\phi$, $z_o = \phi(o)$ and $z_y = \phi(y)$.
We utilize mT5~\citep{xue2021mt5} for encoding texts.
The alignment reward can be the cosine similarity between two embeddings, $\cos(z_o, z_y)$.
However, it is maximized when the generated output $o$ is identical to $y$, ignoring the question language.
To address this, we introduce negative samples by shuffling both outputs and ground-truth responses with the same permutation, $\tilde{z}_o = \phi(\psi(o))$ and $\tilde{z}_y = \phi(\psi(y))$, where $\psi$ denotes the shuffle function.
Inspired by~\citet{schroff2015facenet}, the multilingual context alignment maximizes similarity between positive samples and minimizes similarity between negative samples, enforcing a margin, $\alpha$, between these similarities:
\begin{equation}
    \cos(\tilde{z}_o, \tilde{z}_y) + \alpha < \cos(z_o, z_y).
\end{equation}

The final context alignment reward is defined as:
\begin{equation}
    r_\text{context-align} = \max(\cos(z_o, z_y) - \cos(\tilde{z}_o, \tilde{z}_y) + \alpha, 0),
    \label{eq:context}
\end{equation}
where $\alpha$ denotes the margin, set to $1$, the maximum possible value of cosine similarity.

\paragraph{Multilingual Reasoning-Step Alignment.}
We further introduce a multilingual reasoning-step alignment to provide finer-grained matching. 
Given the split output sentences, $\mathbf{o} = (o^{(1)}, \ldots, o^{(N)})$ and ground-truth sentences $\mathbf{y} = (y^{(1)}, \ldots, y^{(M)})$, each output sentence, $o^{(i)}$ is aligned with a ground-truth sentence, $y^{(j_i)}$.
Since the number of output and reference sentences ($N$ and $M$) may differ, we use dynamic programming to maximize the total similarity between the pairs while preserving order:
\begin{align}
    \max_{1 \leq j_1 \leq ... \leq j_N \leq M} \sum_{i=1}^{N} \mathbf{C}_{i, j_i},
\end{align}

where $\mathbf{C}\in\mathbb{R}^{N\times M}$ is the similarity matrix, and $\mathbf{C}_{i,j}$ denotes the similarity score between embeddings $z_o^{(i)}$ and $z_y^{(j)}$.  
We use the same function used in Eq.~(\ref{eq:context}) for $\mathbf{C}$.
The multilingual reasoning-step alignment reward is then defined as the average similarity across aligned pairs:  
\begin{align}
    r_\text{step-algin} &= \frac{1}{N}\sum_{i=1}^N \mathbf{C}_{i, j_i} = \frac{1}{N}\sum_{i=1}^N \max(\cos(z_o^{(i)}, z_y^{(j_i)}) - \cos(\tilde{z}_o^{(i)}, \tilde{z}_y^{(j_i)}) + \alpha, 0).
    \label{eq:step}
\end{align}

\paragraph{Language Consistency.}
We also define a language consistency reward for giving a more direct incentive to reason in the question language.
Given the output $o$, and language detector $f$~(\eg, Google Translate, langid~\citep{lui2012langid}), the language consistency reward is defined as $1$ if the detected language in response matches the target language $l$, $0$ if it does not:
\begin{align}
    r_\text{lang} = \delta[f(o) = l_t],
\end{align}
where $\delta[\cdot]$ denotes the indicator function.
The final reward is defined as the sum of individual reward:
\begin{align}
    r = r_\text{context-align} + r_\text{step-align} + r_\text{lang}.
\end{align}

%% file: sections/4_benchmark.tex
%!TEX root = ./../main.tex

\begin{figure}[t!]
    \centering
    \includegraphics[width=\columnwidth]{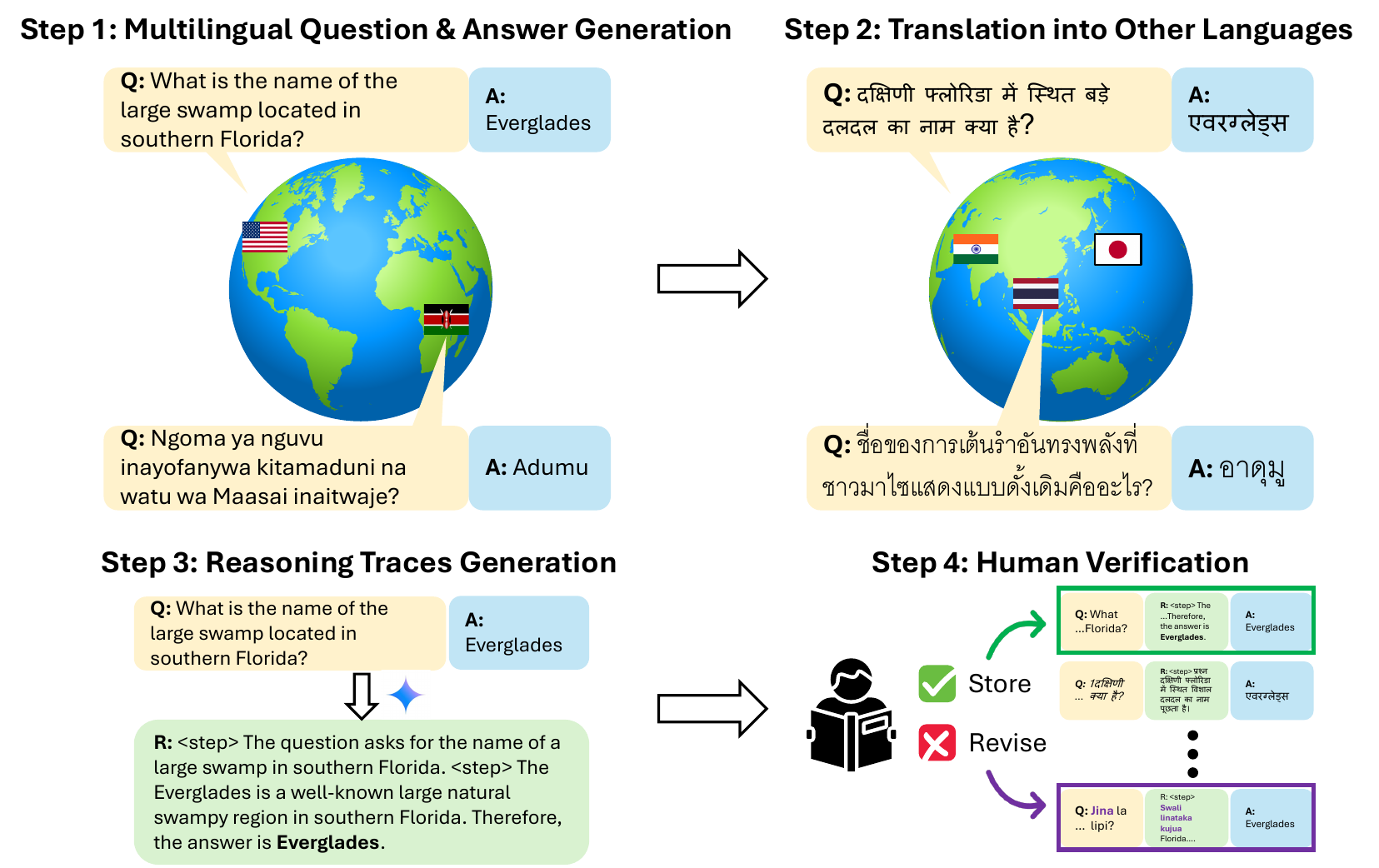}
    \caption{
    \textbf{Illustration of \data benchmark construction.}
    (1) Geography-aware multilingual questions and answers are generated by Gemini 2.0 Flash.
    (2) The data is translated into other languages, verifying whether it is back-translatable.
    (3) The reasoning trace for each question and answer pair is generated.
    (4) Native or C1-level speakers verify each data and revise it if needed.
    }
    \label{fig:data_collection}
\end{figure}

\begin{table}[t!]
\caption{
Comparison between existing multilingual factual or common-sense benchmark and \data.
}
\label{tab:geofact_comparison}
\centering
\resizebox{0.95\textwidth}{!}{%
\setlength{\tabcolsep}{4pt} % Set column separation to 15pt
\begin{tabular}{l|ccccccc}
\toprule
Benchmark & Size & \#Lang. & Geo-Aware & Train Set & Reasoning Eval.\\
\midrule
XStoryCloze~\citep{lin2022few} & 1872 & 11 & & \checkmark\\
XWINO~\citep{tikhonov2021s} & 3961 & 6 & \\
XCOPA\citep{ponti2020xcopa} & 6600 & 11 & & \checkmark (English only)\\
X-FaKT\citep{aggarwal2025language} & 2362 & 13 & \\
XLQA~\citep{roh2025xlqa} & 3000 & 8 & \checkmark \\
\midrule
\data (ours) & 12780 & 5 & \checkmark  & \checkmark & \checkmark\\
\bottomrule
\end{tabular}
}
\vspace{0.2cm}
\end{table}

\section{\data: Geography-Based Factual Reasoning Benchmark}\label{sec:fact_reason_data}

Despite advances in multilingual LLMs \citep{ahuja2023mega,qin2025survey}, robust evaluation of factual reasoning across cultures remains underexplored.
We introduce \data, a benchmark of 3,000 culturally grounded questions (about 600 per country) spanning history, politics, geography, art, and culture, localized to the USA, India, Japan, Kenya, and Thailand in their predominant languages (English, Hindi, Japanese, Swahili, and Thai).
%The goal is to assess factual accuracy and reasoning within culturally specific knowledge spaces.
Our goal is to capture country-specific factual knowledge, encouraging language models to reason effectively within culturally contextualized knowledge spaces.
Table~\ref{tab:geofact_comparison} compares \data with existing multilingual factual reasoning benchmarks.
Our geography-aware multilingual benchmark has a training set and reasoning evaluations compared to other benchmarks.

\begin{figure}[t!]
    \centering
    \includegraphics[width=0.9\textwidth]{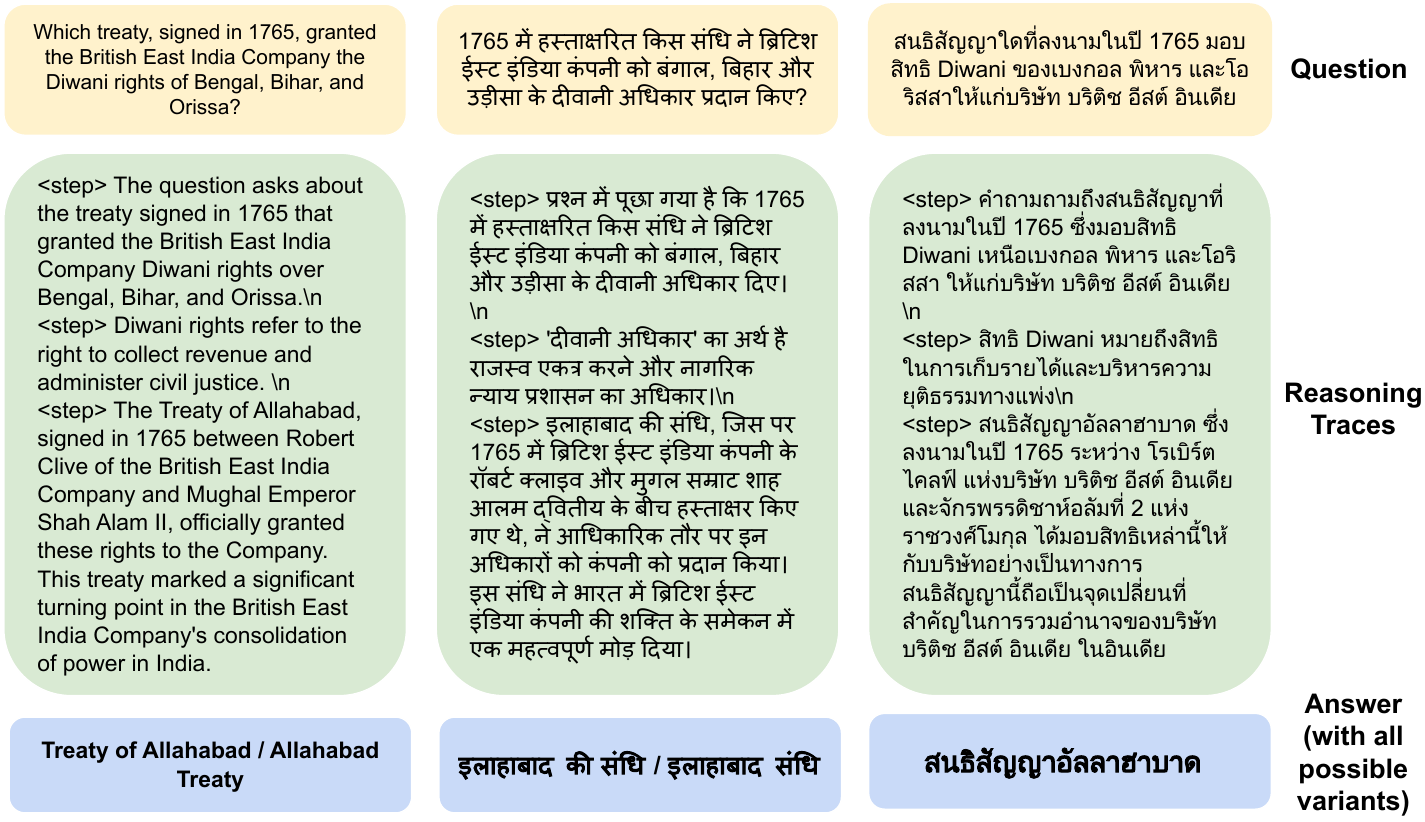}
    \caption{
    \textbf{A sample from \data in English, Hindi, and Thai.}
    Each presents the same factual question and answer content translated across languages.
    These multilingual and semantically equivalent traces serve as reference reasoning for benchmarking the reasoning quality of other language models in our evaluation framework.}
    \label{fig:fact_dataset}
    \vspace{-0.3cm}
\end{figure}

Figure~\ref{fig:data_collection} illustrates the process of the dataset construction.
We adopt a two-stage validation pipeline to ensure factual accuracy and dataset quality.
Rule-based filters and cross-language checks remove incorrect or inconsistent pairs.
Specifically, we verify cross-language answer consistency by translating each answer into English via the Google Translate API to identify mismatches.
Gemini~2.0 Flash~\citep{team2023gemini} then generates structured chain-of-thought reasoning traces for each item, enhancing interpretability and providing supervision signals.
We split the dataset into a train set (85\%) and a test set (15\%), ensuring no semantic overlap across splits, even across languages.
All test samples are manually verified by native or C1-level speakers for factual correctness and linguistic clarity.
Figure~\ref{fig:fact_dataset} illustrates an example multilingual question with its reasoning trace and final answer.

For evaluation, we measure answer accuracy and reasoning score.
Answer accuracy is computed by comparing predictions against the reference answers.
The reasoning score is assessed by Qwen-2.5-72B-Instruct~\citep{qwen2025qwen2.5} as an LLM-as-a-judge, comparing model-generated reasoning traces against the human-verified reasoning traces in the test set.
If a reasoning trace is produced in a language different from the question, identified by a language detector~\citep{lui2012langid}, its score is set to zero. 
Detailed curation, distribution, and evaluation procedures are provided in Appendix~\ref{supp:dataset}.

%% file: sections/5_revisit.tex
%!TEX root = ./../main.tex

\section{Revisiting Multilingual Mathematical Reasoning Benchmark}\label{sec:revisit}

We investigate whether strong performance on multilingual reasoning benchmarks reliably reflects reasoning in the question language.
As a case study, we use MGSM~\citep{shi2023language}, which evaluates multilingual mathematical reasoning in ten diverse languages and provides chain-of-thought prompts (\textsf{Naive-CoT}) in each language to enforce reasoning in the language. 
MGSM reports only mathematical accuracy, implicitly assuming that high accuracy implies language-consistent reasoning.

To address this, we introduce \emph{language accuracy}, which measures whether the generated reasoning matches the intended question language.
Formally, given the language identifier~(\eg, Google Translate, langid~\citep{lui2012langid}), $f$, language accuracy, $A_\text{lang}$ is defined as follows:
\begin{equation}
    A_\text{lang} = \frac{1}{N}\sum_n^N \delta [f(o_n) = l_n],
\end{equation}
where $N$ denotes the number of samples in the dataset, and $\delta[\cdot]$ is indicator function.
$o_n$, and $l_n$ mean the generated output and the target question language, respectively.
Then, we defined the joint accuracy of mathematics and language, $A_\text{joint}$ as follows:
%"
\begin{equation}
    A_\text{joint} = \frac{1}{N} \sum_n^N \left( \delta[f(o_n) = l_n] \cdot \delta[\hat{a}_n = a_n] \right),
\end{equation}
where $\hat{a}_n$ and $a_n$ indicate predicted and ground-truth answers for $n$-th sample, respectively.

\begin{figure}[t!]
    \centering 
    \includegraphics[width=0.7\linewidth]{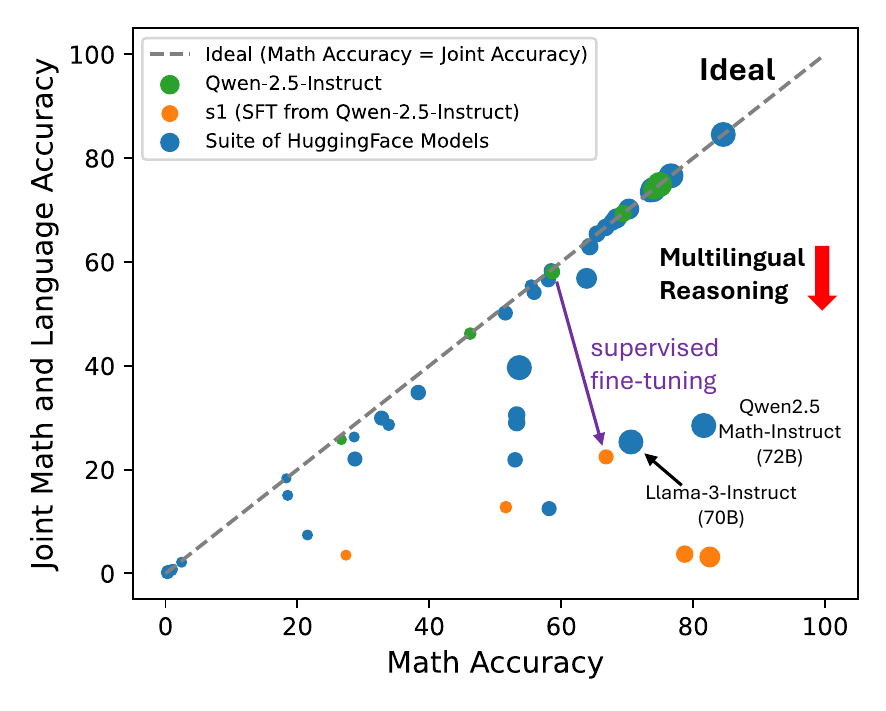}
    \caption{
    \textbf{
    Mathematical accuracy and the joint accuracy of mathematics and language of various LLMs on MGSM with native Chain-of-Thought.}
    Circle size is proportional to the number of parameters.
    The $y=x$ line represents ideal performance, where a model always uses the target question language in reasoning.
    Many models, especially the supervised fine-tuned model, s1, fall significantly below this line, indicating they solve the problem correctly but fail to adhere to the language prompt.
    }
    \label{fig:mgsm}
\end{figure}

We evaluate various recent large language models, including Qwen2.5~\citep{hui2024qwen2}, Llama3~\citep{grattafiori2024llama}, Gemma3~\citep{team2025gemma}, and DeepSeek-R1~\citep{guo2025deepseek}, on MGSM~(see Appendix~\ref{supp:revisit_full_list} for the full list).
Figure~\ref{fig:mgsm} illustrates average mathematical accuracy against joint accuracy across different languages.
Ideally, both metrics should be the same (grey dashed line), yet models such as Qwen2.5-72B-Math-Instruct and Llama-3-70B-Instruct show large gaps, indicating frequent reasoning in the wrong language.
Moreover, the s1 models (orange), fine-tuned from Qwen2.5-Instruct (green), notably degrade language accuracy while improving mathematical performance.
These results demonstrate that mathematical accuracy alone overestimates multilingual reasoning ability, and joint evaluation is essential for assessing true language-consistent reasoning.

%% file: sections/6_exp.tex
%!TEX root = ./../main.tex

\section{Experiments}
\label{sec:exp}

We use Qwen-2.5-7B-Instruct as the backbone for all experiments on mathematical and factual reasoning.
Training and evaluation are conducted on 4 NVIDIA A100 GPUs with DeepSpeed~\citep{rasley2020deepspeed}.
We use three random seeds to calculate the mean and standard error.
Please refer to Appendix~\ref{supp:exp_details} and the attached codebase for the implementation and training details.

\subsection{Dataset}
\label{subsec:exp_details} 

\textbf{Mathematical Reasoning.}
The s1K-1.1 dataset~\citep{muennighoff2025s1} contains 1,000 curated math questions with chain-of-thought traces, selected for difficulty, diversity, and quality. To test multilingual generalization, we additionally construct \textsc{s1K-X}, a multilingual version of s1K-1.1 obtained by translating into ten typologically diverse languages via Google Translate, used for baseline SFT results.  
For evaluation, we report results on GSM8K~\citep{cobbe2021training} and its multilingual counterpart MGSM~\citep{shi2023language} with Native CoT prompts.
We also evaluate language accuracy introduced in Section~\ref{sec:revisit}.

\textbf{Factual Reasoning.}
We utilize \data (Section~\ref{sec:fact_reason_data}), which contains culturally grounded factual QA pairs across five countries (USA, India, Japan, Kenya, Thailand) in five local languages (English, Hindi, Japanese, Swahili, Thai).
Models are trained on the train split and evaluated on the test split.

\begin{table}[t!]
    \caption{
    Accuracy of Qwen2.5-7B-Instruct and post-training methods in GSM8K (English) and MGSM (ten languages).
    Results are reported for mathematical accuracy (Math.), language accuracy (Lang.), and joint accuracy (Joint).
    Bold indicates the best performance in each column.
    }
\label{tab:math}
    \centering
\resizebox{0.97\textwidth}{!}{%
    \begin{tabular}{l|ccc|ccc}
    	\toprule
        \multirow{2}{*}{Method} & \multicolumn{3}{c|}{GSM8K}  & \multicolumn{3}{c}{MGSM}\\
        & Math. & Lang. & Joint & Math. & Lang. & Joint\\
        \midrule
        Qwen-2.5-Instruct & 81.2 & \textbf{100} & 81.2 & 58.7 & 99.0 & 58.1 \\
        \midrule
        GRPO & 80.4 $\pm$ 0.9 & \textbf{100.0 $\pm$ 0.0} & 80.4 $\pm$ 0.9 & 58.8 $\pm$ 0.4 & 95.9 $\pm$ 2.9 & \textbf{58.2 $\pm$ 0.7}\\
        SFT (s1) & 87.2 $\pm$ 1.6 & \textbf{100.0 $\pm$ 0.0} & 87.2 $\pm$ 1.6 & \textbf{66.7 $\pm$ 0.1} & 31.0 $\pm$ 0.5 & 21.9 $\pm$ 0.6\\
        SFT on s1K-X & 84.3 $\pm$ 1.1 & 66.7 $\pm$ 33.3 & 56.5 $\pm$ 28.3 & 45.2 $\pm$ 4.1 & \textbf{99.7 $\pm$ 0.1} & 45.0 $\pm$ 4.3\\
        \model (ours)  & \textbf{87.3 $\pm$ 0.1} & \textbf{100.0 $\pm$ 0.0} & \textbf{87.3 $\pm$ 0.1} & 59.0 $\pm$ 0.3 & 97.8  $\pm$ 0.2 & 58.1 $\pm$ 0.4 \\
        \bottomrule
    \end{tabular}
    }
\end{table}

\subsection{Mathematical Reasoning} 
\label{subsec:math_results}
Table~\ref{tab:math} presents the performance of the base model, Qwen2.5-7B-Instruct, and models with post-training methods, supervised fine-tuning (SFT), GRPO, and \model.
Supervised fine-tuning on s1K-1.1 improves mathematical reasoning performance on GSM8K and MGSM but substantially degrades multilingual performance in MGSM, leading to lower joint accuracy.
Training on the translated multilingual dataset (s1K-X) preserves language accuracy on MGSM but reduces mathematical accuracy.
GRPO, in contrast, produces little change, likely due to sparse rewards.
For instance, Figure~\ref{fig:mgsm_response} shows that GRPO outputs are identical to the base model, whereas SFT produces an English response to a Russian query.

\model outperforms baselines in all metrics on GSM8K and achieves large gains in joint accuracy on MGSM compared to SFT.
Unlike SFT, it preserves reasoning in the query language while still improving mathematical correctness.
In effect, \model learns mathematical reasoning without sacrificing multilingual fidelity, whereas other methods either fail to learn reasoning (GRPO) or lose multilingualism (SFT).
Appendix~\ref{supp:bridge_one_lang} further examines a variant of \model trained with translation into a single language instead of multiple languages, and detailed per-language results are provided in Appendix~\ref{supp:mgsm_indiv}.

\begin{table}[t!]
\caption{Comparison of model performance on average reasoning score~(\%), language accuracy~(\%), and answer accuracy~(\%) on \data test set, evaluated across all examples and split by whether the language is associated with the country (`Assoc.') or not (`Non-Assoc.').
Bold means the best performance. % \paul{too few baselines pls add more.}
}
\label{tab:geofact_main}
\centering
\resizebox{0.97\textwidth}{!}{%
\begin{tabular}{l|ccc|ccc}
\toprule
\multirow{2}{*}{Model} & \multicolumn{3}{c|}{Average Reasoning Score (\%)} & \multicolumn{3}{c}{Average Answer Accuracy (\%)} \\
& All & Assoc. & Non-Assoc. & All & Assoc. & Non-Assoc. \\
\midrule
\small{DeepSeek-R1-Distill-Llama-8B} & 23.9 & 25.7 & 23.4  & 26.3 & 30.5 & 25.2 \\
\t{DeepSeek-R1-Distill-Qwen-7B} & 32.7 & 34.4 & 32.3 &  22.9 & 22.6 & 22.9 \\
Command R7B &  43.1 & 47.5 & 41.9  &  25.8 & 33.7 & 23.8 \\
\midrule
Qwen-2.5-Insturct & 46.5 & 49.5 & 45.8 &  26.2 & 33.7 & 24.3\\
\midrule
GRPO  & 45.6 $\pm$ 0.3 & 48.1 $\pm$ 0.8 & 44.9 $\pm$ 0.2 & 32.1 $\pm$ 0.3 & 37.6 $\pm$ 0.2 & \textbf{30.7 $\pm$  0.3} \\
SFT & 47.7 $\pm$ 0.1 & 50.7 $\pm$ 0.5 & 46.9 $\pm$ 0.1 & 29.3 $\pm$ 0.2 & 37.1 $\pm$ 1.0 & 27.3 $\pm$ 0.3 \\
\model (ours)  & 48.5 $\pm$ 0.4 & 52.6 $\pm$ 0.5 & 47.5 $\pm$ 0.3 & 32.0 $\pm$ 0.6 & \textbf{41.3 $\pm$ 1.0} & 29.7 $\pm$ 0.5 \\
\model (ours, Thai only) & \textbf{50.1 $\pm$ 0.1} & \textbf{53.2 $\pm$ 0.3} & \textbf{49.3 $\pm$ 0.2} & \textbf{32.2 $\pm$ 0.4} & 39.9 $\pm$ 0.4 & 30.2 $\pm$ 0.3\\
\bottomrule
\end{tabular}
}
\end{table}

\subsection{Factual Reasoning}
\label{subsec:fact_results}

Table~\ref{tab:geofact_main} summarizes the performance of the base model (Qwen‑2.5‑Instruct) and the gains obtained after post‑training with GRPO, supervised fine‑tuning (SFT), and \model on \data.
For comparison, we also illustrate the performance of other pretrained LLMs~\citep{cohere2025commandaenterprisereadylarge,guo2025deepseek}.
We report reasoning score and answer accuracy.
Results are additionally split by whether the language is associated with the country (\textit{assoc.}) or not (\textit{non‑assoc.}); for instance, Thai is associated with Thailand but not with the USA.
All pretrained models perform better in associative settings, likely because pretraining corpora contain more paired examples where language and country co-occur.
This gap underscores the challenge of aligning reasoning across languages and contexts, motivating methods that explicitly enforce language consistency.

\model achieves the strongest reasoning performance compared to both pretrained and post-trained baselines, and it is reinforced when only using Thai for translation.
Notably, \model improves both settings at a similar rate (4–6\% in reasoning score and 20–21\% in answer accuracy).
A per-language and per-country breakdown is provided in Appendix~\ref{supp:geofact_each}.  
Figure~\ref{fig:qualitative_lewis} further illustrates model outputs: although all systems reason in the question language (Swahili), only \model predicts the correct answer.

Finally, we apply machine translation as a post-hoc strategy. Table~\ref{tab:geofact_translated} shows that translation via Google Translate offers no substantial improvements over the direct setting (Table~\ref{tab:geofact_main}), reflecting their weaker multilingual alignment.
This suggests that post-hoc translation provides, at most, a superficial fix and fails to address the core challenge of multilingual reasoning.

\begin{table}[t!]
\caption{Machine-translated performance of each model on \data test set.
Google Translate is used to translate the generated output into the question language.
Bold means the best performance.
}
\label{tab:geofact_translated}
\centering
\resizebox{0.97\textwidth}{!}{%
\begin{tabular}{l|ccc|ccc}
\toprule
\multirow{2}{*}{Model with Machine Translation} & \multicolumn{3}{c|}{Average Reasoning Score (\%)} & \multicolumn{3}{c}{Average Answer Accuracy (\%)} \\
& All & Assoc. & Non-Assoc. & All & Assoc. & Non-Assoc. \\
\midrule
DeepSeek-R1-Distill-Llama-8B & 27.6 & 29.2 & 27.2 & 7.9 & 11.1 & 7.1 \\
DeepSeek-R1-Distill-Qwen-7B & 33.2 & 33.9 & 33.1 & 8.8 & 11.1 & 8.2 \\
Command R7B & 44.2 & 48.5 & 43.1 & 25.0 & 34.1 & 22.7 \\
\midrule
Qwen-2.5-Insturct &  45.7 & 49.2 & 44.8 & 28.9 & 36.4 & 27.0 \\
\midrule
GRPO &   45.7 $\pm$ 0.3 & 48.3 $\pm$ 0.3 & 45.1 $\pm$ 0.3 & \textbf{31.9 $\pm$ 0.3} & 37.6 $\pm$ 0.8 & \textbf{30.5 $\pm$ 0.2} \\
SFT & 47.8  $\pm$ 0.1& 50.9 $\pm$ 0.4 & 47.1 $\pm$ 0.1 & 27.2 $\pm$ 0.1 & 35.8  $\pm$ 1.1& 25.0  $\pm$ 0.3\\
\model (ours) & 48.7 $\pm$ 0.4 & 52.8 $\pm$ 0.5 & 47.7 $\pm$ 0.4 & 31.8 $\pm$ 1.0 & \textbf{41.3 $\pm$ 1.2} & 29.4 $\pm$ 0.9 \\
\model (ours, Thai only) & \textbf{50.1 $\pm$ 0.2} & \textbf{53.1 $\pm$ 0.3} & \textbf{49.4 $\pm$ 0.3} & 30.6 $\pm$ 0.8 & 39.1 $\pm$ 0.3 & 28.5 $\pm$ 0.9 \\
\bottomrule
\end{tabular}
}
\end{table}

\begin{table}[t!]
\caption{Contribution of individual reward functions to \model. The evaluation is performed on GSM8K and MGSM.
Bold means the best performance.
Lang: Language Consistency, CA: Context Alignment, RA: Reasoning-Step Alignment.
}
\label{tab:ablation}
\centering
\resizebox{0.97\textwidth}{!}{%
    \begin{tabular}{ccc|ccc|ccc}
    \toprule
 \multicolumn{3}{c|}{M2A Variants}  & \multicolumn{3}{c|}{GSM8K}  & \multicolumn{3}{c}{MGSM}\\
    Lang & CA & RA & Math. & Lang. & Joint & Math. & Lang. & Joint\\
    \midrule
    $\checkmark$ & & & 86.9 $\pm$ 0.0 & \textbf{100.0 $\pm$ 0.0} & 86.9 $\pm$ 0.0 & 54.2 $\pm$ 0.1 & 98.3 $\pm$ 0.1 & 53.8 $\pm$ 0.1 \\
    $\checkmark$ & \checkmark & & 84.7 $\pm$ 0.1 & \textbf{100.0 $\pm$ 0.0}  & 84.7 $\pm$ 0.1 & 57.8 $\pm$ 0.1 & \textbf{99.5 $\pm$ 0.1} & 57.5 $\pm$ 0.1 \\
    $\checkmark$ & \checkmark & \checkmark & \textbf{87.3 $\pm$ 0.1} & \textbf{100.0 $\pm$ 0.0} & \textbf{87.3 $\pm$ 0.1} & \textbf{59.0 $\pm$ 0.3} & 97.8  $\pm$ 0.2 & \textbf{58.1 $\pm$ 0.4} \\ 
\bottomrule
\end{tabular}
}
\end{table}

\subsection{Ablation Study}
\paragraph{Contribution of Individual Reward Functions.}
We analyze the effectiveness of individual reward functions in \model on the mathematical reasoning task.
Table~\ref{tab:ablation} shows that context alignment (CA) improves multilingual performance on MGSM but slightly lowers GSM8K accuracy, as enforcing global embedding similarity adds constraints unnecessary for English-only tasks.
Reasoning-step alignment (RA) provides finer supervision by aligning individual reasoning steps, which boosts multilingual performance and mitigates the small degradation from CA.  
The full model, combining language consistency, CA, and RA, achieves the best results on both benchmarks, confirming that the reward functions are complementary: CA promotes global cross-lingual alignment, while RA enforces stepwise reasoning fidelity.

\paragraph{Reward Formulations.}\label{supp:reward_form}
We compare different formulations of the multilingual alignment reward used in Eq.~(\ref{eq:context}) and Eq.~(\ref{eq:step}).  
Table~\ref{tab:reward_form} reports results on GSM8K and MGSM. 
Using vanilla cosine similarity yields weaker performance, while adding a negative-sample term improves MGSM but slightly reduces GSM8K.
Our margin-based hinge formulation achieves the best results across~all metrics, demonstrating the benefit of combining negative samples with a margin to stabilize alignment.

\begin{table}[t!]
\caption{Comparison of reward formulation for multilingual alignment rewards of \model. The evaluation is performed on GSM8K and MGSM.
Bold means the best performance.
}
\label{tab:reward_form}
\centering
\resizebox{0.97\textwidth}{!}{%
    \begin{tabular}{l|ccc|ccc}
    \toprule
 \multirow{2}{*}{Reward Formulation} & \multicolumn{3}{c|}{GSM8K}  & \multicolumn{3}{c}{MGSM}\\
     & Math. & Lang. & Joint & Math. & Lang. & Joint\\
    \midrule
    $\cos(z_o, z_y)$ & 84.1 $\pm$ 0.1 & 100.0 $\pm$ 0.0 & 84.1 $\pm$ 0.1 &57.4 $\pm$ 0.6 & 97.4 $\pm$ 0.1 & 56.0 $\pm$ 0.2 \\ 
    $\cos(z_o, z_y) - \cos(\tilde{z}_o, \tilde{z}_y)$ & 83.6 $\pm$ 0.1 & 100.0 $\pm$ 0.0 & 83.6 $\pm$ 0.1 & 57.6 $\pm$ 0.1 & \textbf{99.6 $\pm$ 0.1} & 57.4 $\pm$ 0.1 \\
    $\max(\cos(z_o, z_y) - \cos(\tilde{z}_o, \tilde{z}_y) + \alpha, 0)$ (ours) & \textbf{87.3 $\pm$ 0.1} & \textbf{100.0 $\pm$ 0.0} & \textbf{87.3 $\pm$ 0.1} & \textbf{59.0 $\pm$ 0.3} & 97.8  $\pm$ 0.2 & \textbf{58.1 $\pm$ 0.4} \\ 
\bottomrule
\end{tabular}
}
\end{table}

%% file: sections/7_conclusion.tex
%!TEX root = ./../main.tex

\section{Discussion}\label{sec:conclusion}
\vspace{-0.1cm}
 
 We conducted a comprehensive study of whether large language models (LLMs) reason in the language of the input question.
 Our findings show that many LLMs predominantly reason in English or Chinese, even when prompted in other languages, undermining multilingual reasoning quality and limiting their applicability in culturally and linguistically diverse settings.
 
To overcome this limitation, we introduce a novel method, \model, which enforces language-consistent reasoning while preserving factual correctness.
By combining multi-scale alignment rewards with a language-consistency objective, \model aligns outputs with ground-truth reasoning traces at both context and reasoning step levels, encouraging reasoning to remain in the query language.

Robust evaluation of multilingual reasoning is itself difficult, since most benchmarks focus on final answers rather than reasoning quality or language alignment.
We therefore propose \data, a geography-based factual reasoning benchmark spanning five diverse languages, paired with step-by-step reasoning traces and a reasoning evaluation protocol including logical structure, factual correctness, and language consistency.

Our results show that \model consistently improves multilingual mathematical and factual reasoning capability while maintaining strong English performance.
While our experiments were conducted on 7B-parameter models, the approach is scalable and provides a practical alternative to massive multilingual instruction tuning.
More broadly, our contributions establish a foundation for training and evaluating LLMs that reason faithfully across languages, advancing the goal of globally inclusive, culturally grounded, and interpretable AI.

%% file: sections/8_appendix.tex
%!TEX root = ./../main.tex

\begin{figure}[t!]
    \centering
    \includegraphics[width=0.95\textwidth]{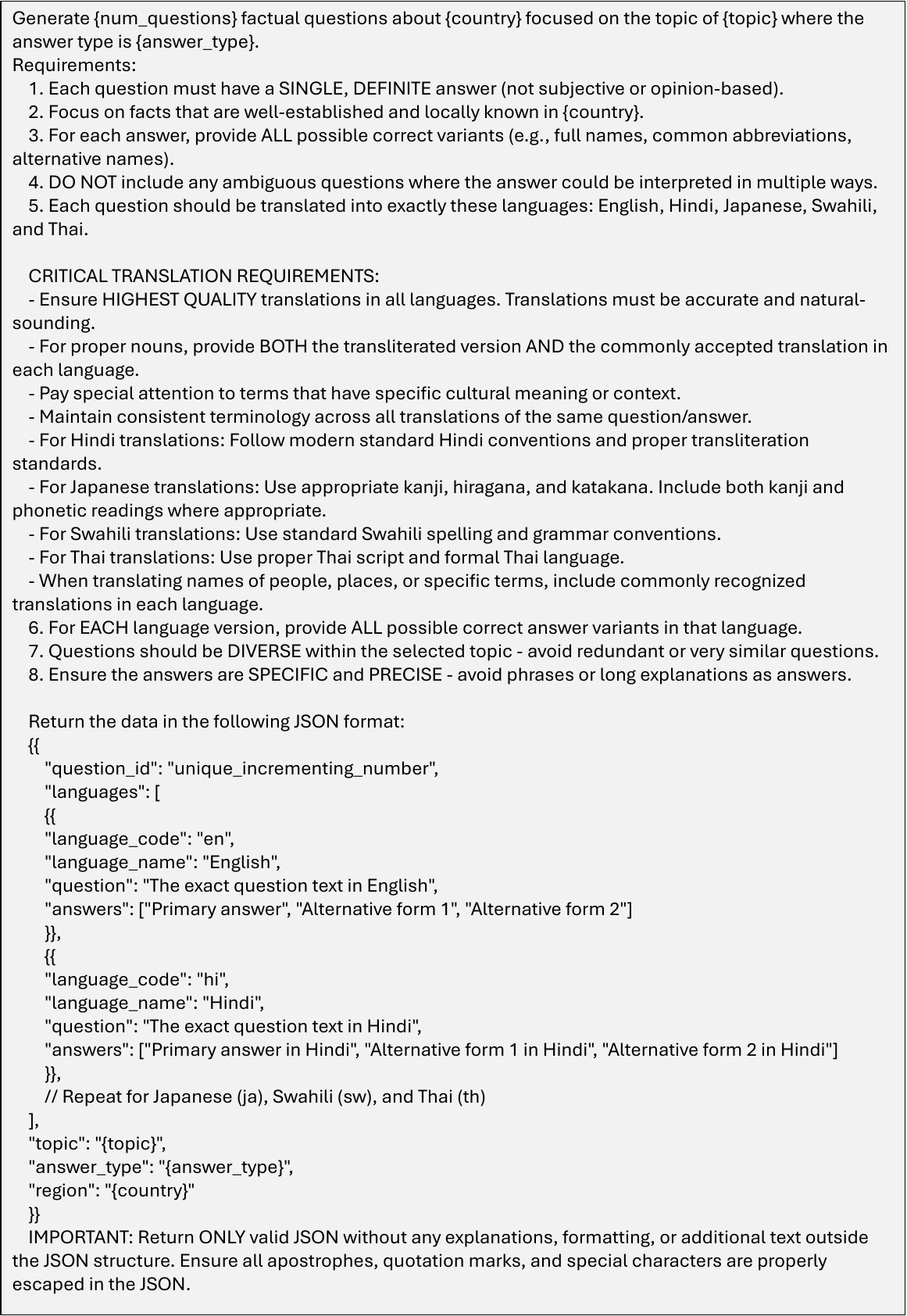}
    \vspace{0.1cm}
    \caption{
    \textbf{Prompt for Generating Multilingual Factual Questions and Answers in \data.}
This prompt instructs the LLM to generate diverse, unambiguous factual questions about a specific country and topic, each with a single, definite answer. The questions and their answers are provided in five languages, English, Hindi, Japanese, Swahili, and Thai, with strict requirements for high-quality translations, consistent terminology, and inclusion of all valid answer variants in each language.}
    \label{fig:llm_generate}
\end{figure}

\section{Details of \data}\label{supp:dataset}

\begin{figure}[t!]
    \centering
    \includegraphics[width=0.75\textwidth]{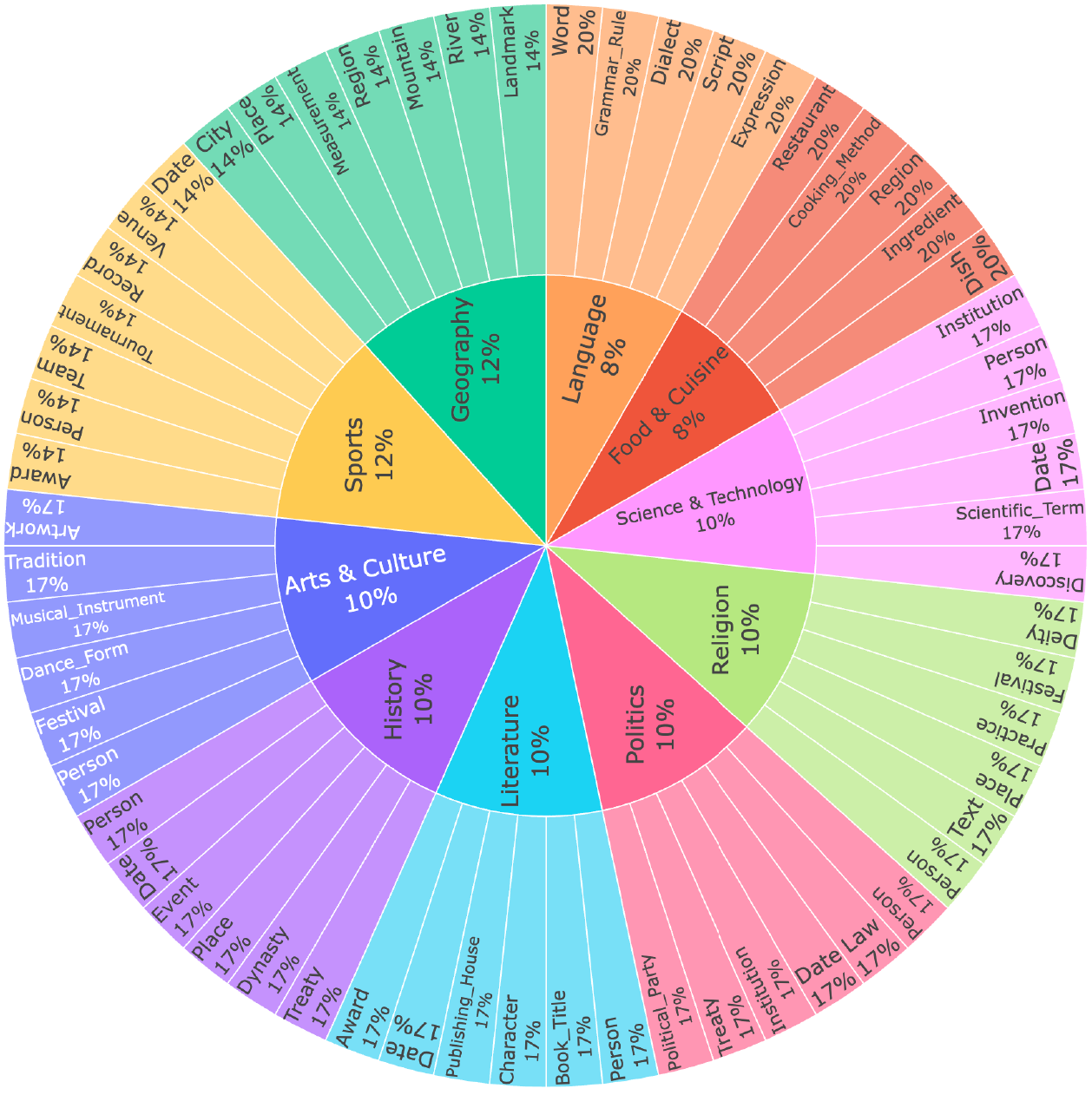}
    \caption{
    \textbf{Illustration of the hierarchical distribution of generated factual question categories by topic and subcategory.}
     Each colored wedge represents a major topic (e.g., History, Geography), and its outer segments represent specific subcategories (e.g., Person, Place, Treaty). The size of each segment reflects the proportion of questions allocated to that subcategory within its topic. This generation schema was applied uniformly across five countries, and all question sets were translated into five different languages.
    }
    \label{fig:pie_chart}
\end{figure}

\subsection{Dataset Collection}
\label{supp:dataset_collection}

\begin{figure}[t!]
    \centering
    \includegraphics[width=0.95\textwidth]{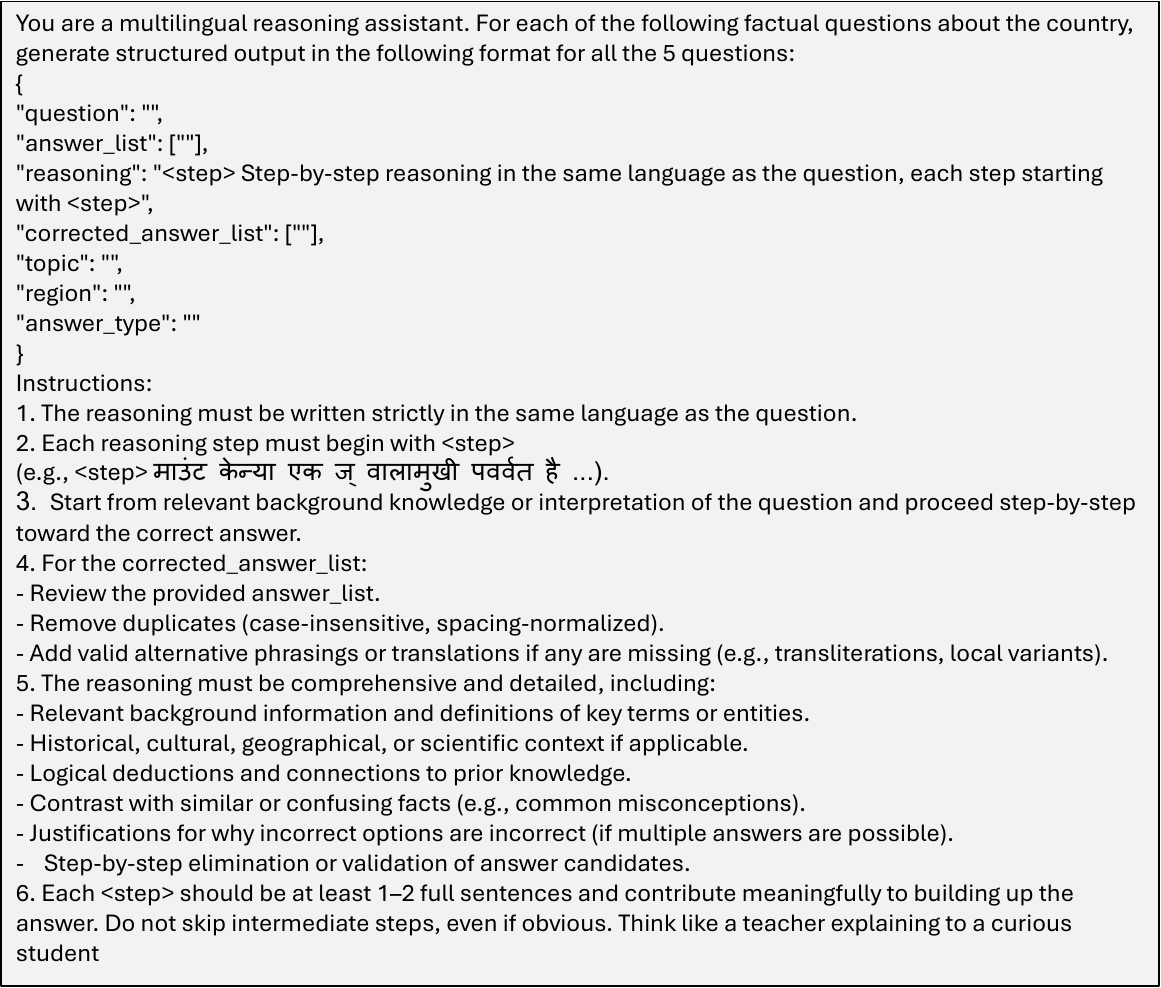}
    \vspace{0.1cm}
    \caption{
\textbf{Structured prompt for multilingual factual reasoning generation using Gemini 2.0 Flash on \data.} This prompt guides the model to generate step-by-step reasoning and corrected answers for factual questions about a country, using the same language as the input question. The output consists of five JSON object strings for the same factual question, each in a different language.
}
    \label{fig:llm_reason}
\end{figure}

We constructed a multilingual factual QA dataset using Gemini~2.0 Flash.
For each country–language pair (USA–English, India–Hindi, Japan–Japanese, Kenya–Swahili, Thailand–Thai), we generated 600 unique QA pairs (3,000 examples in total) by using prompt templates shown in Figure~\ref{fig:llm_generate}.
The topics spanned ten high-level domains: History, Geography, Politics, Literature, Arts \& Culture, Science \& Technology, Sports, Food \& Cuisine, Language, and Religion with subcategories such as \textit{Person}, \textit{Date}, and \textit{Place} (Figure~\ref{fig:pie_chart}).
 For each subcategory, 20 questions were generated per country.
Translations were produced with Google Translate, and semantic fidelity was checked via back-translation.
Reasoning traces were generated by Gemini~2.0 Flash using Chain-of-Thought prompting (Fig.~\ref{fig:llm_reason}), with each step explicitly tagged by a `\textsf{<step>}' token.  

The dataset is split into training (85\%) and test (15\%) sets, with no semantic overlap across splits or languages.
Ten percent of the training data and all test data were manually verified by the authors through cross-referencing with Wikipedia and Google Search.
In addition, all test samples were reviewed by native or C1-level speakers to ensure factual correctness and linguistic clarity and modify the samples if needed.

\begin{figure}[t!]
    \centering
    \includegraphics[width=0.95\textwidth]{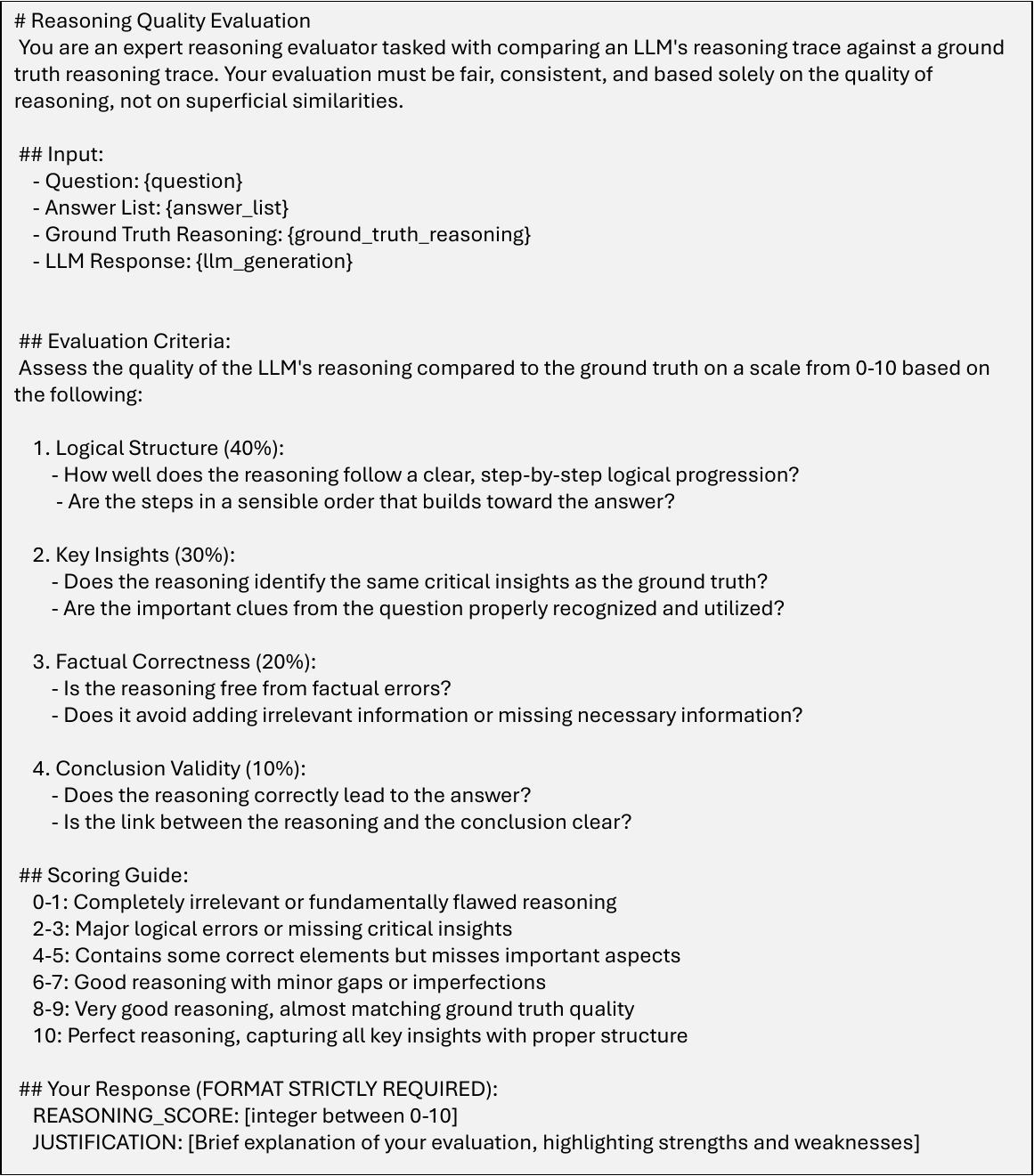}
    \caption{
    \textbf{Prompt for LLM-as-a-Judge to Evaluate Reasoning Traces Using Gemini 2.0 Flash.}
This prompt guides the evaluation of an LLM-generated reasoning trace against a ground truth using specific criteria such as logical structure, key insights, factual correctness, and conclusion validity. The evaluation is performed by Qwen2.5-72B Instruct, acting as the LLM-as-judge, and includes scoring, language mismatch detection, and answer validation.}
    \label{fig:llm_judge}
\end{figure}

\subsection{Evaluation Protocol}\label{supp:geofact_evaluation}
The benchmark has three metrics, answer accuracy, and reasoning score.
Answer accuracy is computed by checking whether the model prediction appears in the list of reference answers provided for each test instance.  
Reasoning score is evaluated with Qwen-2.5-72B-Instruct~\citep{qwen2025qwen2.5} as an LLM-as-a-Judge, which compares model-generated reasoning traces against human-verified references to measure the quality of generated reasoning.
 If a reasoning trace is written in a different language from the question, detected by a language identifier~\citep{lui2012langid}, its score is set to zero.
 Figure~\ref{fig:llm_judge} illustrates the prompt structure used for the LLM-as-a-Judge, including the evaluation instructions and rules applied to model outputs.

\subsection{License}
We release the dataset under the MIT License, which permits reuse, modification, and distribution, provided that the original license and copyright notice are included.

\section{Experimental Details}\label{supp:exp_details}

\subsection{Training}
We adopt the hyperparameter configuration from s1~\citep{muennighoff2025s1}, with the exception of batch size, which we reduced from 16 to 8 due to resource constraints. Specifically, the training hyperparameters are as follows: learning rate of $10^{-5}$, minimum learning rate of $0$, weight decay of $10^{-4}$, total batch size of $8$, training conducted for $5$ epochs, and a cosine learning rate scheduler with warmup ratio of $0.05$.
Adam~\citep{kingma2015adam} is used with $\beta_1 = 0.9$ and $\beta_2 = 0.95$.
The maximum sequence and token lengths are set to 20,000.
GRPO uses accuracy and format rewards following~\citet{guo2025deepseek}.
\model uses a maximum completion length of 1,024 ($256$) tokens, generates 2~($8$) completions per prompt, and sets the maximum step to $10$ due to the resource constraints (parentheses denote factual reasoning parameters in case of difference).
We also use a loss coefficient for GRPO as $0.01$ for mathematical reasoning and $0.5$ for factual reasoning.
We train the model with three different random seeds to calculate the standard error.
For the s1K-X dataset, we use Google Translate to translate the s1K-1.1 dataset into multiple languages used in the MGSM benchmark: Bengali, German, Spanish, French, Japanese, Russian, Swahili, Telugu, Thai, and Chinese.

\subsection{Evaluation}
\label{supp:evaluation}

For mathematical reasoning, we employed the \textsf{lm-evaluation-harness}\footnote{\url{https://github.com/EleutherAI/lm-evaluation-harness}} library to evaluate each model.
 Specifically, we used the MGSM~\citep{shi2023language} \textsf{Native-CoT} setting and the MMLU-ProX~\citep{xuan2025mmluprox} Math category with a 5-shot chain-of-thought prompt to ensure the model reasons in its native language.
langid~\citep{lui2012langid} is used to evaluate language correctness.

\subsection{Models Evaluated on MGSM}
\label{supp:revisit_full_list}

Table~\ref{tab:full_model_list} provides the complete list of models evaluated in Figure~\ref{fig:mgsm}.
All models are sourced from HuggingFace\footnote{\url{https://huggingface.co/}}, a public repository of large language models.
We include Qwen2.5~\citep{qwen2025qwen2.5}, s1~\citep{muennighoff2025s1}, Llama~\citep{grattafiori2024llama}, Gemma~\citep{team2025gemma}, and DeepSeek-R1~\citep{guo2025deepseek}, each with a range of model sizes.

\begin{table}[t!]
\caption{List of all models and sizes evaluated on MGSM in Figure~\ref{fig:mgsm}. All models are sourced from HuggingFace.
}
\vspace{0.1cm}
\centering
\resizebox{0.6\textwidth}{!}{%
\begin{tabular}{l|ccc}
\toprule 
Model Name &  Model Sizes\\
\midrule
Qwen2.5 & 1.5B, 3B, 7B, 14B, 32B\\
Qwen2.5-Instruct & 1.5B, 3B, 7B, 14B, 32B\\
Qwen2.5-Instruct-GPTQ-Int4 & 1.5B, 3B, 7B, 14B, 32B\\
Qwen2.5-Instruct-GPTQ-Int8 & 1.5B, 3B, 7B, 14B, 32B\\
Qwen2.5-Instruct-AWQ & 7B, 14B, 32B\\
Qwen2.5-Instruct-1M & 7B, 14B\\
Qwen2.5-Math & 1.5B, 7B, 72B\\
Qwen2.5-Math-Instruct & 1.5B, 7B, 72B\\
\midrule
s1 & 1.5B, 3B, 7B, 14B, 32B\\
\midrule
Llama-3-Instruct & 8B, 70B\\
Llama-3.3-Instruct & 70B\\
\midrule
Gemma-3-PT & 1B, 4B, 12B, 27B\\
Gemma-3-IT & 1B, 4B, 12B, 27B\\
\midrule
DeepSeek-R1-Distill-Qwen & 1.5B, 3B, 7B, 14B, 32B \\
DeepSeek-R1-Distill-Llama & 8B, 70B \\
\bottomrule
\end{tabular}
}
\label{tab:full_model_list}
\end{table}

\section{\model with Different Languages}\label{supp:bridge_one_lang}
In the main paper, we present \model trained with random translations drawn from the ten MGSM languages (Bengali, German, Spanish, French, Japanese, Russian, Swahili, Telugu, Thai, and Chinese).
Here, we examine the effect of using a single fixed translation language, as shown in Table~\ref{tab:diff_lang}.
We select Japanese and Swahili as representative examples of high- and low-resource languages, respectively, following the categorization of~\citet{nicholas2023lost}.

Across both choices, we observe only minor decreases in GSM8K mathematical accuracy and MGSM language accuracy relative to the multi-language setting.
This finding indicates that training with a single language—even a low-resource one—can still induce strong multilingual reasoning ability.
Nevertheless, randomizing translations across multiple languages yields the strongest overall results, suggesting that language diversity provides additional regularization benefits for cross-lingual alignment.

\begin{table}[t!]
    \caption{
    Accuracy of \model trained with various languages.
    All languages denote that the language translator randomly translates the question language into ten different languages used in MGSM.
    Bold denotes the best performance for each metric.
    \vspace{0.2cm}
    }
\label{tab:diff_lang}
    \centering
\resizebox{0.9\textwidth}{!}{%
    \begin{tabular}{l|ccc|ccc}
    	\toprule
        \multirow{2}{*}{\shortstack{Translation\\Language}} & \multicolumn{3}{c|}{GSM8K}  & \multicolumn{3}{c}{MGSM}\\
        & Math. & Lang. & Joint & Math. & Lang. & Joint \\
        \midrule
        All & \textbf{87.3 $\pm$ 0.1} & \textbf{100.0 $\pm$ 0.0} & \textbf{87.3 $\pm$ 0.1} & \textbf{59.0 $\pm$ 0.3} & \textbf{97.8  $\pm$ 0.2} & \textbf{58.1 $\pm$ 0.4} \\
        Japanese & 86.8 $\pm$ 0.1 & 100.0 $\pm$ 0.0 & 86.8 $\pm$ 0.1 & 59.0 $\pm$ 0.2 & 90.7 $\pm$ 0.4 & 56.5 $\pm$ 0.1\\ 
        Swahili &  85.3 $\pm$ 0.7 & 100.0 $\pm$ 0.0 & 85.3 $\pm$ 0.7 & 58.4 $\pm$ 0.3 & 96.2 $\pm$ 1.0 & 56.6 $\pm$ 0.7 \\
        \bottomrule
    \end{tabular}
    }
\end{table}

\input{tables/mgsm}

\section{MGSM Evaluation in Each Language}\label{supp:mgsm_indiv}

Table~\ref{tab:mgsm_indiv} shows individual math and language accuracy change compared to the base model (Qwen2.5-7B-Instruct) in each language.
As we mentioned in Section~\ref{sec:exp}, supervised fine-tuning on s1K-1.1 improves math performance while losing language performance.
Conversely, GRPO rarely changes performance on both mathematics and language across all languages.
\model generally maintains language performance.
However, its German performance is much lower, which might be related to the language sampled for translating the question.

\begin{figure}[t!]
    \centering
    \includegraphics[width=\textwidth]{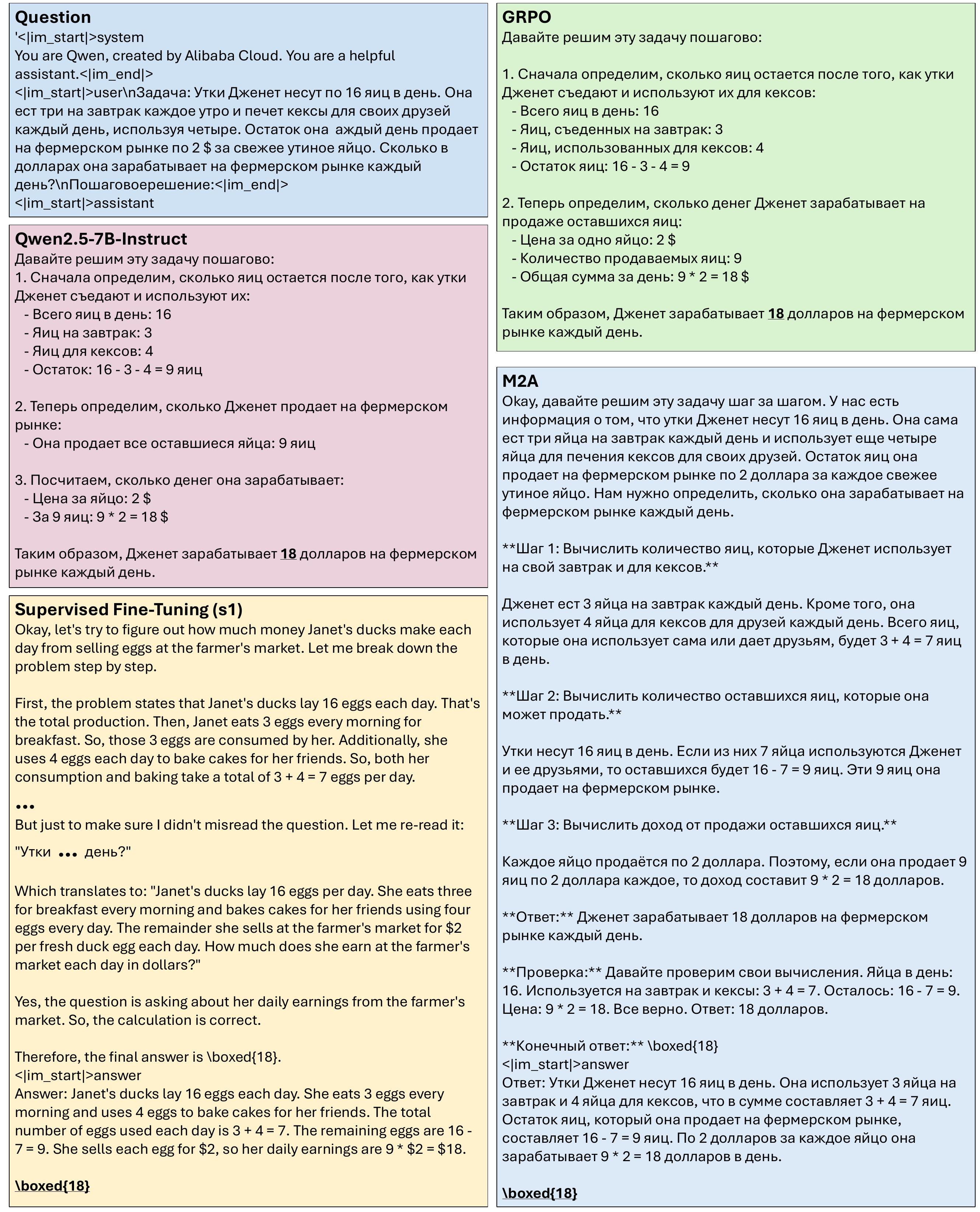}
    \vspace{-0.25cm}
    \caption{
    \textbf{Example question and response of each model on MGSM Russian.}
    The question in English is ``Janet's ducks lay sixteen eggs per day. She eats three for breakfast every morning and bakes muffins for her friends every day with four. She sells the remainder at the farmers' market daily for two dollars per fresh duck egg. How much in dollars does she make every day at the farmers' market?''
    All models correctly provide the answer ($18$), but the SFT model conducts its reasoning in English.
    GRPO's reasoning process is almost identical to that of the base model (Qwen2.5-7B-Instruct).
    }
    \label{fig:mgsm_response}
    \vspace{-0.3cm}
\end{figure}

We also illustrate an MGSM Russian example response from Qwen2.5-7B-Instruct and SFT, GRPO, \model fine-tuned models on Figure~\ref{fig:mgsm_response}.
All models generate the correct answer from a given question written in Russian.
However, the SFT model uses English instead of Russian, while others reason in Russian.
GRPO has an almost identical reasoning process to the base model, which might explain why it performs almost the same in any metrics as the base model.

\input{tables/geofactx_reasoningscore}
\input{tables/geofactx_answer}

\section{\data Evaluation in Each Language and Country Pair}
\label{supp:geofact_each}

Tables~\ref{tab:reasoning_score}, and~\ref{tab:answer_correctness} present reasoning score, language accuracy, and answer accuracy for each language–country pair.  
Diagonal entries correspond to associative pairs, while off-diagonal entries correspond to non-associative pairs.  
Associative countries consistently achieve higher performance across all metrics compared to non-associative ones.
English (a high-resource language) generally outperforms other languages.  
Among non-associative cases, performance for the United States exceeds that of Kenya and Thailand, which may reflect differences in pre-training and fine-tuning data exposure.
As discussed in Section~\ref{subsec:fact_results}, all methods substantially increase language accuracy as the training dataset includes all five languages.
However, no consistent trend is observed in reasoning score and answer accuracy across language–country pairs.

\begin{figure}[t!]
    \centering
    \includegraphics[width=0.95\textwidth]{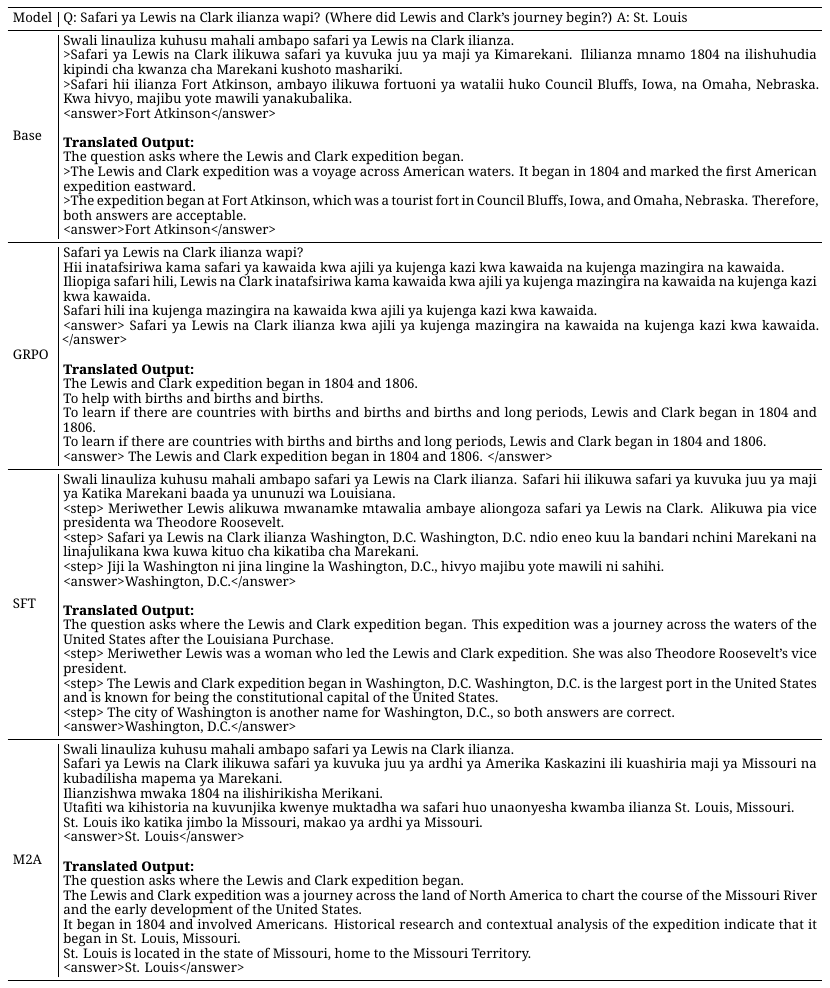}
    \vspace{-0.1cm}
        \caption{
        \textbf{Generated outputs from a given question written in Swahili on \data.}
    All models use Swahili, but only \model generates the correct answer, St Louis.
    }
    \label{fig:qualitative_lewis}
\end{figure}

%% file: tables/mgsm.tex
%!TEX root = ./../main.tex

\begin{table}[t!]
\caption{
Accuracy of base (Qwen2.5-7B-Instruct) model and models fine-tuned with each post-training method on MGSM.
Standard error is not included for readability.
Bold means the best performance.
}
\label{tab:mgsm_indiv}
    \centering
\resizebox{\textwidth}{!}{%
    \begin{tabular}{c|c|cccccccccc}
    	\toprule
        \multirow{2}{*}{Model} &  \multirow{2}{*}{MGSM} & \multicolumn{10}{c}{Question Language}\\ \cline{3-12}
        && bn & de & es & fr & ja & ru & sw & te & th & zh \\ \midrule
        \multicolumn{12}{c}{Math Performance}\\ \midrule
        %%%%%%%%% 7B %%%%%%%%%%%
Qwen2.5-Instruct&58.7&61.2&72.0&72.8&62.4&70.4&65.6&14.0&29.6&69.6&69.2\\
\midrule
GRPO             & 58.8 & 59.7 & 69.7 & 75.9 & 65.7 & 66.9 & 71.7 & 12.8 & 27.7 & 64.9 & 72.4 \\
SFT (s1)         & \textbf{66.7} & \textbf{66.4} & \textbf{77.4} & \textbf{78.0} & \textbf{79.6} & \textbf{73.6} & \textbf{83.6} & \textbf{18.6} & \textbf{35.2} & \textbf{74.6} & \textbf{79.8} \\
SFT on s1K-X     & 45.2 & 34.7 & 51.2 & 60.8 & 69.3 & 44.0 & 52.5 & 10.4 & 8.8  & 64.7 & 55.1 \\
\model           & 59.0 & 53.3 & 75.6 & 75.2 & 75.7 & 66.3 & 80.5 & 3.7 & 14.9& 66.5 & 78.3\\
%\model (JA only) & 64.9 & \textbf{67.1} & 77.2 & 80.4 & \textbf{82.4} & 70.5 & \textbf{84.8} & 16.1 & 27.1 & 67.2 & 75.6 \\
%\model (RU only) & 62.6 & 59.5 & 74.8 & 79.9 & 80.4 & 64.8 & 78.0 & 13.6 & 25.7 & 68.1 & \textbf{80.8} \\
%\model (SW only) & 58.4 $\pm$ 0.3 & 51.3 $\pm$ 2.0 & 75.3 $\pm$ 0.1 & 76.9 $\pm$ 1.5 & 77.9 $\pm$ 1.3 & 62.9 $\pm$ 2.1 & 82.4 $\pm$ 0.6 & 7.6 $\pm$ 1.0 & 16.9 $\pm$ 1.4 & 58.9 $\pm$ 1.2 & 74.1 $\pm$ 0.9 \\
\midrule
        \multicolumn{12}{c}{Language Performance} \\ \midrule
Qwen2.5-Instruct&99.0&\textbf{100.0}&\textbf{100.0}&\textbf{100.0}&\textbf{100.0}&91.6&\textbf{100.0}&\textbf{100.0}&\textbf{100.0}&98.8&\textbf{100.0}\\
\midrule
GRPO             & 95.9 & 99.9  & \textbf{100.0} & \textbf{100.0} & 99.9  & \textbf{100.0} & \textbf{100.0} & 60.0  & \textbf{100.0} & 99.6  & \textbf{100.0} \\
SFT (s1)         & 31.0 & 13.6  & 81.4  & 88.6  & 5.8   & 6.0   & 2.6   & 0.0   & 22.8  & 27.8  & 62.2 \\
SFT on s1K-X     & \textbf{99.7} & \textbf{100.0} & 99.9  & 99.9  & 99.9  & \textbf{100.0} & 99.7  & 99.9  & \textbf{100.0} & \textbf{100.0} & 98.0 \\
\model           & 97.8 & 99.7 & 98.0 & 98.5 & 94.8 & 99.9 & 99.2 & 88.3& 99.9 & 99.3 & \textbf{100.0} \\
%\model (JA only) & 64.6 & 23.2  & 98.7  & 78.3  & 87.1  & 98.9  & 31.7  & 4.9   & 56.8  & 66.8  & 99.6 \\
%\model (RU only) & 78.6 & 69.9  & 66.0  & 66.9  & 64.3  & 96.9 & 99.6  & 61.7  & 93.3  & 67.7  & 99.9 \\
%\model (SW only) & 96.2 $\pm$ 1.0 & 93.2 $\pm$ 6.8 & 99.1 $\pm$ 0.4 & 98.8 $\pm$ 0.2 & 96.3 $\pm$ 3.5 & 98.9 $\pm$ 1.1 & 94.4 $\pm$ 2.3 & 99.7 $\pm$ 0.3 & 83.5 $\pm$ 12.6 & 98.9 $\pm$ 0.5 & 99.7 $\pm$ 0.3\\
\midrule
        \multicolumn{12}{c}{Joint Performance}\\
        \midrule
Qwen2.5-Instruct&58.1&\textbf{61.2}&72.0&72.8&62.4&65.2&65.6&\textbf{14.0}&\textbf{29.6}&\textbf{68.8}&69.2\\
\midrule
GRPO             & \textbf{58.2} & 59.6 & 69.7 & 75.9 & 65.6 & 66.9 & 71.7 & 7.7  & 27.7 & 64.5 & 72.4 \\
SFT (s1)         & 21.9 & 5.6  & 62.8 & 68.6 & 4.6  & 2.4  & 1.2  & 0.0  & 5.8  & 18.6 & 49.0 \\
SFT on s1K-X     & 45.0 & 34.7 & 51.2 & 60.8 & 69.3 & 44.0 & 52.5 & 10.4 & 8.8  & 64.7 & 53.7 \\
\model           & 58.1 & 53.2 & \textbf{74.1} & \textbf{74.3} & \textbf{71.2} & \textbf{66.3} & \textbf{80.4} & 2.7  & 14.9 & 66.0 & \textbf{78.3} \\
%\model (JA only) & 44.9 & 13.2 & \textbf{76.4} & 62.4 & 71.1 & \textbf{69.7} & 26.5 & 0.7  & 12.9 & 40.9 & 75.3 \\
%\model (RU only) & 48.5 & 39.3 & 47.1 & 51.6 & 51.5 & 62.8 & 78.0 & 7.2  & 23.9 & 43.1 & \textbf{80.7} \\
%\model (SW only) & 56.6 $\pm$ 0.7 & 46.9 $\pm$ 4.9 & 74.8 $\pm$ 0.2 & 76.5 $\pm$ 1.3 & 74.9 $\pm$ 4.1 & 62.1 $\pm$ 2.9 & 78.3 $\pm$ 1.5 & 7.6 $\pm$ 1.0 & 12.5 $\pm$ 2.7 & 58.1 $\pm$ 1.2 & 74.1 $\pm$ 0.9 \\
\bottomrule
    \end{tabular}
    }
\end{table}

%% file: tables/geofactx_reasoningscore.tex
%!TEX root = ./../main.tex
\begin{table}[t!]
\caption{Average reasoning score (\%) by language and region. Reasoning quality is assessed using an LLM-as-a-judge framework, which evaluates model-generated justifications against reference Gemini 2.0 Flash reasoning traces in the \data dataset.
Higher scores indicate more coherent, relevant, and logically sound reasoning.
The gray diagonal entries represent associated language–country pairs.
Bold means the best performance in each pair.
}
\vspace{0.1cm}
\centering
%\scriptsize
\resizebox{0.75\textwidth}{!}{
\begin{tabular}{llccccc}
\toprule
Language & Model & USA & India & Japan & Kenya & Thailand \\
\midrule
\multirow{2}{*}{English} 
  & Qwen2.5-Instruct & \cellcolor[RGB]{240,240,240} 67.5&55.7&\textbf{62.8}& 56.6&51.4  \\
& GRPO & \cellcolor[RGB]{240,240,240}\textbf{70.3 $\pm$ 1.2} & 61.8 $\pm$ 1.3 & 58.5 $\pm$ 1.2 & 57.2 $\pm$ 0.7 & \textbf{51.9 $\pm$ 0.6} \\
  & SFT & \cellcolor[RGB]{240,240,240}69.2 $\pm$ 0.6 & 62.9 $\pm$ 0.7 & 57.6 $\pm$ 0.6 & 57.4 $\pm$ 0.1 & 51.2 $\pm$ 0.7 \\
  & \model & \cellcolor[RGB]{240,240,240}69.3 $\pm$ 0.5 & \textbf{63.0 $\pm$ 0.7} & 57.8 $\pm$ 0.7 & \textbf{57.4 $\pm$ 0.2} & 51.5 $\pm$ 0.8 \\
\hline
\multirow{2}{*}{Hindi} 
  & Qwen2.5-Instruct &  \textbf{39.4}&\cellcolor[RGB]{240,240,240}\textbf{39.0} & \textbf{39.0}&35.0&38.6 \\
  & GRPO  & 38.7 $\pm$ 1.4 & \cellcolor[RGB]{240,240,240}36.6 $\pm$ 0.9 & 38.2 $\pm$ 0.7 & 35.5 $\pm$ 0.7 & 38.2 $\pm$ 0.9 \\
  & SFT & 36.3 $\pm$ 1.0 & \cellcolor[RGB]{240,240,240}36.0 $\pm$ 0.5 & 37.7 $\pm$ 1.4 & 42.4 $\pm$ 0.5 & \textbf{40.8 $\pm$ 0.9} \\
  & \model & 36.5 $\pm$ 1.1 & \cellcolor[RGB]{240,240,240}36.0 $\pm$ 0.4 & 38.2 $\pm$ 1.1 & \textbf{42.5 $\pm$ 0.5} & 40.8 $\pm$ 0.9 \\
\hline
\multirow{2}{*}{Japanese} 
  & Qwen2.5-Instruct &  \textbf{51.4}&43.1&\cellcolor[RGB]{240,240,240} \textbf{53.2} &45.2&40.3 \\
  & GRPO &50.1 $\pm$ 1.4 & \textbf{43.2 $\pm$ 0.6} & \cellcolor[RGB]{240,240,240}52.3 $\pm$ 0.1 & 47.0 $\pm$ 1.2 & 43.5 $\pm$ 0.6 \\
  & SFT & 45.3 $\pm$ 1.1 & 43.0 $\pm$ 0.6 & \cellcolor[RGB]{240,240,240}52.3 $\pm$ 1.3 & 47.7 $\pm$ 0.7 & 43.0 $\pm$ 0.2 \\
  & \model & 45.5 $\pm$ 1.1 & 43.0 $\pm$ 0.5 & \cellcolor[RGB]{240,240,240}52.6 $\pm$ 1.4 & \textbf{48.1 $\pm$ 0.7} & \textbf{43.5 $\pm$ 0.3} \\
\hline
\multirow{2}{*}{Swahili} 
  & Qwen2.5-Instruct &  41.4& 43.3& 39.5& \cellcolor[RGB]{240,240,240}41.3& 34.9 \\
  & GRPO  & 34.6 $\pm$ 0.5 & 33.2 $\pm$ 1.4 & 31.4 $\pm$ 0.6 & \cellcolor[RGB]{240,240,240}35.6 $\pm$ 1.2 & 33.8 $\pm$ 0.4 \\
  & SFT  & 47.9 $\pm$ 0.2 & \textbf{49.8 $\pm$ 1.3} & 47.4 $\pm$ 0.2 & \cellcolor[RGB]{240,240,240}49.6 $\pm$ 0.4 & 47.3 $\pm$ 0.7 \\
  & \model &  \textbf{48.0 $\pm$ 0.3} & \textbf{49.8 $\pm$ 1.3} & \textbf{47.5 $\pm$ 0.2} & \cellcolor[RGB]{240,240,240}\textbf{49.7 $\pm$ 0.4} & \textbf{47.3 $\pm$ 0.8} \\
\hline
\multirow{2}{*}{Thai} 
  & Qwen2.5-Instruct &   \textbf{56.4}& 44.0& \textbf{49.0}& 43.5&\cellcolor[RGB]{240,240,240}  45.3 \\
    & GRPO & 54.1 $\pm$ 1.4 & \textbf{47.5 $\pm$ 0.3} & 46.7 $\pm$ 1.6 & \textbf{47.8 $\pm$ 0.6} & \cellcolor[RGB]{240,240,240}45.0 $\pm$ 0.6 \\
  & SFT & 40.3 $\pm$ 0.8 & 40.2 $\pm$ 0.3 & 47.2 $\pm$ 0.6 & 45.1 $\pm$ 0.1 & \cellcolor[RGB]{240,240,240}45.5 $\pm$ 0.4 \\
  & \model &  40.3 $\pm$ 0.6 & 40.5 $\pm$ 0.3 & 47.5 $\pm$ 0.6 & 45.5 $\pm$ 0.2 & \cellcolor[RGB]{240,240,240}\textbf{45.6 $\pm$ 0.4} \\
\bottomrule
\end{tabular}
}
\label{tab:reasoning_score}
\end{table}

%% file: tables/geofactx_answer.tex
%!TEX root = ./../main.tex

\begin{table}[t!]
\caption{Average answer accuracy by language and region.
The gray diagonal entries represent associated language–country pairs.
Bold means the best performance in each pair.
}
\vspace{0.1cm}
\centering
\resizebox{0.75\textwidth}{!}{
\begin{tabular}{ll|ccccc}
\toprule
Lang & Model & USA & India & Japan & Kenya & Thailand \\
\midrule
  \multirow{2}{*}{English} & Qwen2.5-Instruct & \cellcolor[RGB]{240,240,240}56.3 & 32.7 & 41.9 & 31.6 & 24.4 \\
& GRPO& \cellcolor[RGB]{240,240,240}\textbf{74.7 $\pm$ 1.1} & \textbf{62.9 $\pm$ 2.1} & \textbf{52.0 $\pm$ 2.4} & \textbf{49.1 $\pm$ 2.1} & \textbf{38.4 $\pm$ 0.7} \\
& SFT & \cellcolor[RGB]{240,240,240}70.9 $\pm$ 1.9 & 60.2 $\pm$ 2.0 & 47.0 $\pm$ 1.3 & 45.3 $\pm$ 0.6 & 34.5 $\pm$ 1.0 \\
& \model & \cellcolor[RGB]{240,240,240}65.6 $\pm$ 2.2 & 59.0 $\pm$ 3.0 & 51.7 $\pm$ 2.2 & 43.1 $\pm$ 5.6 & 31.1 $\pm$ 2.3 \\
\hline
\multirow{2}{*}{Hindi} & Qwen2.5-Instruct & 19.5& \cellcolor[RGB]{240,240,240} \textbf{23.3}&\textbf{16.1}&14.3&20.0\\
& GRPO& \textbf{22.8 $\pm$ 1.5} & \cellcolor[RGB]{240,240,240}22.1 $\pm$ 1.8 & 11.9 $\pm$ 2.0 & 13.9 $\pm$ 1.0 & \textbf{20.8 $\pm$ 0.8} \\
& SFT & 11.4 $\pm$ 1.6 & \cellcolor[RGB]{240,240,240}14.7 $\pm$ 2.1 & 10.0 $\pm$ 0.4 & 18.3 $\pm$ 0.4 & 15.0 $\pm$ 1.4 \\
& \model & 15.1 $\pm$ 1.8 & \cellcolor[RGB]{240,240,240}19.7 $\pm$ 1.7 & 13.3 $\pm$ 1.9 & \textbf{19.8 $\pm$ 1.2} & 14.5 $\pm$ 2.4 \\
\hline
\multirow{2}{*}{Japanese} & Qwen2.5-Instruct & \textbf{36.8}&23.0&\cellcolor[RGB]{240,240,240}39.3&20.2&18.9 \\
& GRPO& 34.6 $\pm$ 2.7 & \textbf{25.3 $\pm$ 1.8} & \cellcolor[RGB]{240,240,240}40.5 $\pm$ 2.1 & 29.0 $\pm$ 1.4 & \textbf{26.6 $\pm$ 2.4} \\
& SFT & 32.9 $\pm$ 1.3 & 20.3 $\pm$ 0.8 & \cellcolor[RGB]{240,240,240}42.1 $\pm$ 4.0 & \textbf{29.4 $\pm$ 0.8} & 19.8 $\pm$ 1.2 \\
& \model & 30.8 $\pm$ 2.6 & 25.1 $\pm$ 0.6 & \cellcolor[RGB]{240,240,240}\textbf{42.5 $\pm$ 2.0} &  27.9 $\pm$ 3.1 &  20.3 $\pm$ 2.0 \\
\hline
\multirow{2}{*}{Swahili} & Qwen2.5-Instruct & 24.3&32.6&17.6&\cellcolor[RGB]{240,240,240}22.6&13.5 \\
& GRPO& 29.3 $\pm$ 0.9 & 25.6 $\pm$ 1.8 & 15.5 $\pm$ 0.4 & \cellcolor[RGB]{240,240,240}19.6 $\pm$ 1.7 & 18.9 $\pm$ 0.0 \\
& SFT & 32.4 $\pm$ 2.1 & \textbf{34.5 $\pm$ 2.5} & \textbf{21.7 $\pm$ 0.8} & \cellcolor[RGB]{240,240,240}26.7 $\pm$ 1.4 & \textbf{25.7 $\pm$ 2.1} \\
& \model & \textbf{33.3 $\pm$ 3.5} & 33.3 $\pm$ 1.1 & 21.1 $\pm$ 1.8 & \cellcolor[RGB]{240,240,240}\textbf{28.3 $\pm$ 3.4} & 22.2 $\pm$ 0.7 \\
\hline
\multirow{2}{*}{Thai}    & Qwen2.5-Instruct &   31.9&23.8&18.5&20.7&\cellcolor[RGB]{240,240,240}25.7 \\
& GRPO&  \textbf{39.6 $\pm$ 1.7} & \textbf{29.2 $\pm$ 0.8} & \textbf{29.2 $\pm$ 1.1} & \textbf{30.1 $\pm$ 2.3} & \cellcolor[RGB]{240,240,240}29.3 $\pm$ 3.2 \\
& SFT & 15.0 $\pm$ 0.5 & 19.6 $\pm$ 1.1 & 20.6 $\pm$ 0.4 & 19.5 $\pm$ 0.7 & \cellcolor[RGB]{240,240,240}\textbf{29.3 $\pm$ 0.5} \\
& \model & 28.6 $\pm$ 3.4 & 23.3 $\pm$ 0.0 & 27.8 $\pm$ 3.6 & 28.6 $\pm$ 2.4 & \cellcolor[RGB]{240,240,240}24.3 $\pm$ 1.8 \\
\bottomrule
\end{tabular}
}
\vspace{1mm}
\label{tab:answer_correctness}
\end{table}